\documentclass[a4paper,fleqn]{cas-sc}

\usepackage[numbers,sort&compress]{natbib}

\usepackage{amsmath}
\usepackage{amssymb}
\usepackage{amsthm}

\usepackage{graphicx}
\usepackage[export]{adjustbox}
\usepackage{xcolor}
\usepackage{array}
\usepackage{setspace}
\usepackage[most]{tcolorbox}

\usepackage{caption}
\usepackage{subcaption}
\usepackage{algorithm}
\usepackage[noend]{algpseudocode}

\usepackage{hyperref}
\usepackage{cleveref}

\begin{document}
\doublespacing

\let\WriteBookmarks\relax
\def\floatpagepagefraction{1}
\def\textpagefraction{.001}

\shorttitle{Agentic Robotic Additive Manufacturing Process Planning} 

\shortauthors{Ge et~al.}

\title [mode = title]{Kinematics-Grounded Agentic AI for Robotic Additive Manufacturing Process Planning}

\author[1]{Jingzhan Ge}
\author[1]{Ruimin Chen}
\author[2]{Azadeh Haghighi}
\author[1]{Jiong Tang}
\author[1]{Farhad Imani\cormark[1]}
\ead{farhad.imani@uconn.edu}

\cortext[1]{Corresponding author}

\address[1]{School of Mechanical, Aerospace, and Manufacturing Engineering,
University of Connecticut, Storrs, Connecticut, USA}

\address[2]{Department of Mechanical and Industrial Engineering,
University of Illinois Chicago, Chicago, IL 60608, USA}

\begin{abstract}
Robotic additive manufacturing (AM) extends material-extrusion printing beyond gantry kinematics but makes process planning robot-dependent. A slicer-generated plan that appears favorable in part coordinates can become infeasible or robotically unfavorable on a manipulator because slicer-process decisions and part orientation determine the generated path, while part orientation and workspace placement affect its kinematic realization. Existing AM tools, large language model (LLM)-based decision-support methods, and digital-shadow systems do not provide integrated pre-execution evaluation of these coupled decisions. This paper presents agentic robotic additive manufacturing (A-RAM), an agent--specialist--tool framework that converts user intent and a part file into traceable, execution-ready plans. The LLM interprets manufacturing objectives and constraints, identifies prescribed and searchable planning variables, and encodes this reasoning in a schema-constrained request; a deterministic Planning Agent instantiates the corresponding search workflow, while domain tools compute quantitative evidence for slicing, placement, inverse kinematics, trajectory timing, Joint-6 jerk, and extrusion. The framework is evaluated on a six-axis robotic-arm AM cell through three case studies covering expert-specified planning, goal-only planning, objective-dependent infill screening, and geometry-dependent orientation--placement selection. Across the evaluated candidate sets, selected plans achieve up to 53.5\% lower maximum Joint-6 jerk and 48.3\% lower mean absolute Joint-6 jerk than the least favorable valid candidates, while objective-specific infill screening yields motion-plan completion times up to 40.1\% shorter and extrusion paths up to 12.7\% shorter than the corresponding least favorable screened patterns.

\end{abstract}

\maketitle


\section{Introduction}

Robotic additive manufacturing (AM) extends fabrication beyond the geometric and kinematic constraints of conventional gantry-based systems by exploiting the configurational flexibility of manipulators. The additional degrees of freedom enable large-format fabrication, multi-axis deposition, dynamically reoriented printing, and task-specific configuration of the robot workspace and build platform \cite{bhatt2020expanding, fry2020robotic, barjuei2024digital}. This flexibility, however, fundamentally changes the process-planning problem. In gantry-based printing, the machine realization of a deposition path is largely prescribed by the Cartesian toolpath. In robotic AM, the same deposition path can be executed through multiple joint-space realizations determined by part placement, robot configuration, inverse-kinematics (IK) branch selection, transition-motion generation, and trajectory timing \cite{rescsanski2025towards}. Consequently, a geometrically valid toolpath does not uniquely determine a feasible or favorable robot motion. The resulting joint-space behavior also feeds directly into the deposition process because configuration-dependent tool-center-point (TCP) velocity, acceleration, and motion continuity influence synchronized material extrusion \cite{badarinath2021integration}. Slicer-level decisions, including infill pattern, infill density, infill orientation, and part orientation, therefore do more than modify the deposited geometry. They regenerate waypoint sequences, transition motions, robot configurations, and joint-space motion responses, while build-plate placement changes the IK feasibility and motion quality of otherwise equivalent deposition paths.

Existing studies provide evidence for individual elements of this coupling, but slicer-level process decisions and robot-execution decisions have largely been studied in isolation. Slicer studies demonstrate that infill pattern, density, and angle affect filament consumption, printing time, mechanical response, and strength--time tradeoffs \cite{suteja2021effect, lubombo2018effect, de2023effects, birosz2022effect, kim2022continuously}. These investigations establish that slicer parameters generate materially different deposition and travel paths. Robot-aware studies, in parallel, show that relocating a fixed G-code trajectory changes the available IK solutions and positional deviation \cite{GHUNGRAD2023644, bhatt2021optimizing}, that workpiece placement constrains feasible TCP speed \cite{stradovnik2024workpiece}, and that minimum-time trajectory planning must explicitly account for jerk and motion continuity \cite{huang2018optimal, liu2013time}. However, these studies hold either the slicer-generated path or the robot-execution configuration fixed. They therefore do not provide a unified candidate-level evaluation of how slicer-process decisions change the generated path, extrusion length, and planned completion time, while part orientation and workspace placement change IK feasibility and joint-space motion for the resulting path.

This coupling cannot be reduced to a universal planning heuristic because the preferred solution depends jointly on the manufacturing objective, part geometry, generated path, and robot configuration. Material efficiency is associated with extrusion path length, production speed with trajectory completion time, and motion quality with joint-space behavior; however, these quantities are not interchangeable. A shorter deposited path does not necessarily produce a shorter planned completion time because the timed Cartesian trajectory also includes non-deposition transitions and junction-dependent acceleration and deceleration. Likewise, low Cartesian path variation does not guarantee smooth joint motion because the nonlinear IK mapping can amplify small TCP changes into large joint-space accelerations or jerk. Placement optima are also obtained for a particular part and trajectory \cite{GHUNGRAD2023644, bhatt2021optimizing, ghungrad2024three}; therefore, a placement that is favorable for one geometry, orientation, or infill pattern cannot be assumed to remain favorable for another. Each candidate must be evaluated from the G-code it generates, through the IK and time-parameterized trajectory that realize it, to the execution metrics produced in robot joint space. 

These characteristics motivate an agentic orchestration layer for robotic AM process planning, while execution-critical evaluation remains quantitative and tool-based. Candidate feasibility and ranking are therefore grounded in explicit models of deposition geometry, robot kinematics, IK feasibility, time-parameterized joint motion, joint-space jerk, and synchronized extrusion.

Existing AM-oriented LLM systems support knowledge access, process-parameter reasoning, defect-related prediction, runtime monitoring, and manufacturing workflow coordination \cite{CHANDRASEKHAR2024100232, pak2025additivellm, pak2026additivellm2, JADHAV2025105027, HOLLAND2024100}. However, these systems generally do not generate robot-specific feasibility, trajectory, and joint-motion evidence for comparing candidate deposition plans before execution, while digital-shadow and digital-twin systems primarily provide execution-side observability after key process-planning decisions have already been made \cite{rachmawati2023digital}.

We propose agentic robotic additive manufacturing (A-RAM), an agent--specialist--tool planning framework that converts a natural-language manufacturing objective and a part STL into a traceable, execution-ready robotic AM plan. An LLM-based Triage Agent interprets the manufacturing objective and user constraints, determines which planning variables are prescribed and which require exploration, and encodes this planning intent in a schema-constrained request associated with the supplied part. A deterministic Planning Agent then instantiates the interpreted request as the required search and evaluation workflow, determining how the identified variables are enumerated or conditionally expanded under the available planning rules. Domain specialists coordinate candidate-placement generation, slicer-parameter assignment, G-code generation, batch IK evaluation, robot motion planning, time-parameterized trajectory analysis, joint-jerk computation, extrusion-path evaluation, and execution monitoring. This separation assigns the LLM responsibility for manufacturing-intent interpretation and planning-variable reasoning while keeping execution-critical numerical evidence grounded in deterministic domain tools.

For every candidate, A-RAM preserves the G-code, candidate placement and orientation, feasible or infeasible IK outcomes, planned joint trajectory, trajectory completion time, extrusion length, joint-jerk metrics, and final selection as structured artifacts. The resulting evidence chain makes the decision auditable from the original natural-language objective to the robot motion and extrusion commands delivered to the manufacturing cell. It also permits rejected candidates to be inspected rather than silently discarded, exposing whether rejection results from reachability, IK discontinuity, unfavorable joint motion, excessive execution time, or objective-specific tradeoffs. We instantiate and validate A-RAM on a six-axis robotic-arm AM cell. Tool orientation is held fixed during planar deposition, thereby isolating the coupling among path-generating slicer variables, part orientation, workspace placement, IK, trajectory timing, and joint-space motion. This controlled setting supports full-factorial evaluation of candidate plans together with low-cost physical execution, while retaining the central planning challenges that distinguish robotic AM from conventional gantry-based printing. The main contributions are threefold.

\begin{enumerate}
\item We introduce a kinematics-grounded agent--specialist--tool architecture for robotic AM in which an LLM interprets manufacturing objectives and constraints, reasons over which planning variables are prescribed or require exploration, and encodes this intent in a schema-constrained request; a deterministic Planning Agent instantiates the resulting request as a query-conditioned workflow, and domain tools compute and constrain execution-critical geometric, kinematic, trajectory, joint-motion, and extrusion quantities.

\item We formulate robotic AM planning as a coupled candidate-level evaluation of slicer variables, part orientation, and workspace placement, using kinematics-grounded robot-side evidence that includes IK feasibility, time-parameterized execution, extrusion path length, and joint-space jerk rather than slicer-only estimates or unverified model predictions.

\item We instantiate and validate A-RAM through three case studies on a six-axis robotic-arm AM testbed, demonstrating query-dependent search scoping, objective-dependent tradeoffs, placement-dependent joint-space behavior, and geometry-dependent changes in the selected manufacturing plan.
\end{enumerate}

The remainder of this paper is organized as follows. Section~\ref{sec:related} introduces the related work. Section~\ref{sec:methodology} presents the proposed methodology, kinematics-grounded toolchain, and agent workflow. Section~\ref{sec:results} reports the experimental design, quantitative results, and robotic-cell demonstrations. Finally, Section~\ref{sec:conclusion} summarizes the findings, identifies the current limitations, and discusses future work.

\section{Related Work}
\label{sec:related}

This section reviews four bodies of work underlying robotic AM planning, including agentic tool coordination, slicer variables that generate candidate paths, robot-side execution planning for fixed paths, and monitored execution.

\subsection{Agentic and LLM-based manufacturing decision support}
\label{sec:related_llm}

Agent formulations established that language models can interleave reasoning with external actions and observations, positioning the model as a coordinator of tools rather than only a generator of text \cite{yao2022react}. In AM, one line of work stores domain knowledge in the model or retrieval system. AMGPT combines a pretrained Llama2 model with a curated AM corpus and live literature retrieval for contextual question answering \cite{CHANDRASEKHAR2024100232}. AdditiveLLM predicts defect-related outcomes and performs better with structured parameter inputs than with natural-language prompts \cite{pak2025additivellm}, while AdditiveLLM2 extends adaptation to multimodal inputs with task-dependent accuracy \cite{pak2026additivellm2}. FDM-Bench further shows that G-code anomaly detection and user-query performance vary substantially across model and task type \cite{eslaminia2025fdmbench}. Closer to execution, LLM-3D~Print uses a pretrained vision-language model for print monitoring and corrective reasoning during a running print \cite{JADHAV2025105027}. These systems support AM knowledge access, prediction, and runtime assistance, but they do not close the planning loop from executable AM decisions, such as infill pattern and density, part orientation, and workspace placement, to robot-specific feasibility verification, joint-trajectory evaluation, and synchronized extrusion commands.

A second line constrains the LLM to workflow decomposition and tool coordination. In fiber-composite manufacturing, an LLM agent performs process-chain setup, time estimation, resource allocation, and cycle-time planning \cite{HOLLAND2024100}. Design-to-fabrication demonstrations translate natural-language concepts into meshes and printer-ready G-code \cite{makatura2023llm}. Configuration-selection studies pair a setting generator with an evaluator that computes the objective score for FFF parameters \cite{samani2026programming}. Agentic alloy evaluation uses tool calls to compute thermophysical property diagrams and lack-of-fusion process maps for alloys such as SS316L and IN718 \cite{pak2026agentic}. These works validate tool-grounded orchestration, but their tools operate at the knowledge, process-chain, configuration, design-to-G-code, or materials-property level. They do not verify candidate plans in the workspace of the executing robot. For execution-critical robotic AM planning, the LLM must therefore remain a coordinator, while domain tools produce the numerical evidence used for selection.

\subsection{Slicer variables as path-generating decisions}
\label{sec:related_slicing}

Material-extrusion process planning is multi-objective as path strategies affect accuracy, surface quality, build time, material use, and structural or functional properties \cite{jiang2020path}. At the toolpath level, infill angle adjustment, non-deposition travel reduction, and smoothing of sharp connections affect build time, deposition error, and fabrication quality \cite{jin2014optimization}. The generated path also depends on geometric representation and layer strategy. Direct CAD slicing improves contour accuracy relative to STL-based slicing \cite{han2022additive}, while planar and nonplanar layers differ in support demand, surface quality, build time, dimensional accuracy, and mechanical properties \cite{nayyeri2022planar}. Within a fixed layer strategy, infill pattern, density, and angle affect filament length, print time, mechanical response, energy consumption, and path length \cite{suteja2021effect, lubombo2018effect, de2023effects}. Pattern selection also shifts the strength--time tradeoff \cite{birosz2022effect}, and continuously varied infill can improve tensile performance while reducing print time on unmodified hardware \cite{kim2022continuously}.

These studies establish that slicer settings are not passive profile entries. They generate different extrusion paths, travel motions, waypoint sequences, and process-performance tradeoffs. However, the reported evaluations are performed in part coordinates. In robotic AM, the same slicer decision must also be evaluated after it is mapped into robot joint space. A fixed G-code file is therefore insufficient when infill pattern, density, angle, or part orientation remain open decisions; candidate G-code must be regenerated before robot-side scoring.

\subsection{Robot-side planning for fixed deposition paths}
\label{sec:related_robot}

Robotic material extrusion showed that deposition is not governed by slicer-generated Cartesian commands alone: serial-robot dynamics and pose-dependent TCP-speed variation require coordinated control of robot motion and extrusion \cite{badarinath2021integration}. Nonplanar and conformal robotic AM further couple path generation with reachability, nozzle-orientation planning, collision avoidance, joint-limit enforcement, and continuity constraints \cite{shembekar2019generating, zhao2018nonplanar}. Existing robot-aware methods mainly improve execution after a path has already been generated.

One group smooths or optimizes the robot motion along a fixed path. Time-jerk optimal and jerk-continuous trajectory formulations explicitly account for joint-trajectory smoothness during time optimization \cite{huang2018optimal, liu2013time}. In robotic AM, tool-orientation or tool-axis redundancy can create multiple robot realizations of the same deposition path \cite{shembekar2019generating}. Dai et al.\ select among alternative IK solutions under discrete-time constraints for jerk-optimized redundant-robot trajectories \cite{9025760}. Zhang et al.\ integrate singularity and collision handling and show that discontinuous joint motion near singular regions can produce over- and under-extrusion in deposition \cite{zhang2021singularity}. Chen et al.\ co-optimize tool orientation, kinematic redundancy, and waypoint timing for robot-assisted manufacturing \cite{chen2025co}, while surrogate-guided formulations reduce search cost by predicting joint jerk for nozzle-orientation selection \cite{rescsanski2026constrained}. These methods improve how a robot executes a given path; they do not choose the slicer variables that generate the path.

A second group changes execution by relocating the part. For a prescribed G-code trajectory, workspace location changes IK solutions, energy consumption, and maximum positional deviation \cite{GHUNGRAD2023644}. Placement selection improves deposition accuracy in robot-based AM cells \cite{bhatt2021optimizing}, map-based methods precompute workspace energy-quality distributions for multi-robot placement \cite{ghungrad2024three}, and multi-robot WAAM studies optimize the placement of several manipulators around a common build \cite{bhatt2022optimizing}. Related work in robot machining shows the same dependency: workpiece placement bounds feasible TCP-speed capability along a trajectory and determines whether the robot can execute the path efficiently within joint limits \cite{stradovnik2024workpiece}. These studies show that workspace placement strongly affects robot execution, but they rank placements for a trajectory produced under fixed slicing settings. The joint space of slicer variables, part orientation, and workspace placement remains largely unsearched.

\subsection{Monitored execution and digital shadows}
\label{sec:related_shadow}

Digital manufacturing systems differ by data flow; a digital model has no automatic data exchange, a digital shadow receives one-way physical-to-digital updates, and a digital twin requires bidirectional coupling \cite{kritzinger2018digital}. AM digital-twin frameworks organize execution around data acquisition, model development, calibration, validation, and real-time monitoring \cite{ben2024digital, chen2023multisensor}. Implemented systems demonstrate this monitoring role. A WAAM digital shadow fuses sensor streams to detect bead-level and geometry-related defects during deposition \cite{mu2024digital}, and a robotic wire-based laser metal deposition cell integrates robot state, process data, and sensor feedback for in-situ quality monitoring \cite{alvares2025digital}. Recent architecture-level work also connects LLMs with digital twins under Industry 5.0 \cite{chen2025integrating}.

These systems improve observability after a process plan exists. They do not produce the pre-execution evidence used to choose among candidate slicer, placement, and robot-motion plans. For the planning problem considered here, monitored execution is therefore a traceability layer where live robot and extrusion states are compared against a selected plan whose feasibility and motion quality have already been verified.

Taken together, existing AM LLM systems support knowledge access, prediction, recommendation, monitoring, or process-chain orchestration; slicer studies evaluate path-generating variables in part coordinates; robot-aware planners improve fixed paths or fixed-G-code placements; and digital shadows observe execution after planning. What remains missing is a tool-grounded planning loop that generates candidate deposition jobs and verifies them before execution across slicer variables, part orientation, workspace placement, IK feasibility, trajectory timing, extrusion length, and joint-space motion quality. Such a planning loop requires a clear separation among language-level planning reasoning, deterministic workflow instantiation, and execution-critical computation.


\section{Methodology}
\label{sec:methodology}

A-RAM is formulated as a query-conditioned, tool-grounded planning framework for robotic additive manufacturing. An LLM-based Triage Agent interprets the natural-language manufacturing objective and constraints and identifies the planning variables that are prescribed or require exploration; a deterministic Planning Agent instantiates this structured intent as a constrained search and evaluation workflow; and deterministic manufacturing and robot-execution tools evaluate the resulting candidate deposition plans. The framework returns an execution-ready plan whose numerical evidence is traceable from slicing to monitored execution.

\begin{figure}
    \centering
    \includegraphics[width=1\linewidth]{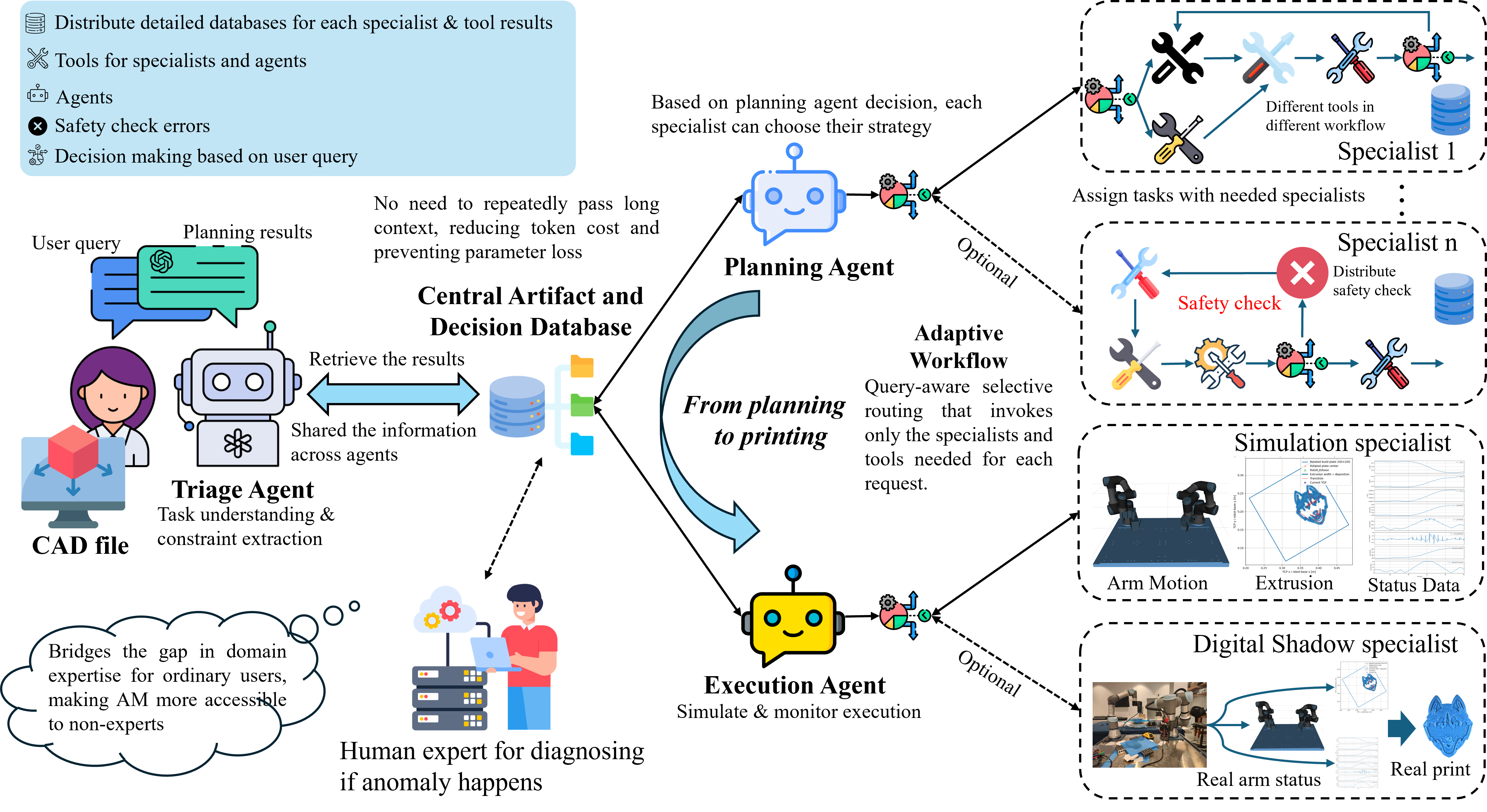}
    \caption{Overview of A-RAM from natural-language manufacturing intent to tool-grounded candidate evaluation and monitored execution. Candidate artifacts are retained throughout the workflow to provide traceability from user intent to G-code, robot motion, and extrusion commands.}
    \label{fig:A-RAM_overview}
\end{figure}

\subsection{Problem formulation}
\label{sec:method_problem}

Let $u$ denote the user request, $\mathcal{M}$ the input part mesh, and $\mathcal{R}$ the calibrated robotic AM cell. The planning decision is a candidate deposition job $x=(p,\theta,\mathbf{c})$, where $p\in\mathcal{P}$ denotes a slicer-process candidate, including the infill pattern and slicer settings, $\theta\in\Theta$ denotes the in-plane part orientation, and $\mathbf{c}\in\mathbb{R}^{2}$ denotes the build-plate placement center. After geometric screening, the nominal product space is restricted by the part geometry and build-plate limits, giving
\begin{equation}
\mathcal{X}(\mathcal{M})
=
\left\{
(p,\theta,\mathbf{c})
\,\middle|\,
p\in\mathcal{P},\;
\theta\in\Theta(\mathcal{M}),\;
\mathbf{c}\in\mathcal{C}(\mathcal{M},\theta)
\right\}.
\label{eq:candidate_space}
\end{equation}

The user request does not necessarily activate the full candidate space. The LLM-interpreted planning request identifies which variables are prescribed by the user and which remain open for exploration, and the Planning Agent deterministically instantiates these decisions as a query-specific search scope $\sigma(u)\subseteq\mathcal{X}(\mathcal{M})$. Expert-constrained queries therefore evaluate only the variables left open by the user, whereas goal-only queries expand the search over slicer variables, part orientation, workspace placement, or their combinations.

A screening evaluation can use a G-code or numerical evaluation profile that differs from the final print profile. Let $\eta\in\mathcal{H}$ denote this profile and let $\eta_{\mathrm{F}}$ denote the final, user-specified print profile. The profile records only evaluation choices, for example, a zero-infill screening override or a trajectory sampling interval, and is not an additional user decision variable. An evaluated job is therefore the tagged tuple $z=(x,\eta)$. Each evaluated job is processed through the deterministic chain
\begin{equation}
(x,\eta)
\;\xrightarrow{\;\mathcal{G}\;}\;
g(x;\eta)
\;\xrightarrow{\;\mathcal{T}\;}\;
\boldsymbol{\tau}(x;\eta)
\;\xrightarrow{\;\mathcal{K}\;}\;
\mathbf{q}(t;x,\eta)
\;\xrightarrow{\;\mathcal{A}\;}\;
\mathbf{m}(x;\eta).
\label{eq:evaluation_chain}
\end{equation}

Here, $\mathcal{G}$ generates the transformed STL and G-code, $\mathcal{T}$ converts the G-code into a time-parameterized Cartesian trajectory, $\mathcal{K}$ solves the sequential inverse-kinematics problem and produces the joint trajectory, and $\mathcal{A}$ extracts the candidate evidence vector
\begin{equation}
\mathbf{m}(x;\eta)=
\left[
\chi_{\mathrm{bed}}(x;\eta),\,
\chi_{\mathrm{slice}}(x;\eta),\,
\chi_{\mathrm{ik}}(x;\eta),\,
\chi_{\mathrm{joint}}(x;\eta),\,
t_{\mathrm{f}}(x;\eta),\,
\ell_{\mathrm{e}}(x;\eta),\,
j_{6}^{\max}(x;\eta),\,
\bar{j}_{6}(x;\eta)
\right]^{\top}.
\label{eq:metric_vector}
\end{equation}

The binary variables $\chi_{\mathrm{bed}}$, $\chi_{\mathrm{slice}}$, $\chi_{\mathrm{ik}}$, and $\chi_{\mathrm{joint}}$ indicate build-plate fit, slicer success, inverse-kinematics feasibility, and joint-continuity feasibility, respectively. The scalar $t_{\mathrm{f}}$ is the trajectory end time, $\ell_{\mathrm{e}}$ is the extrusion path length, and $j_{6}^{\max}$ and $\bar{j}_{6}$ are the maximum and mean absolute Joint-6 jerk values, respectively. For a fixed evaluation profile $\eta$, the feasible candidate set is
\begin{equation}
\mathcal{X}_{\mathrm{f}}(u;\eta)
=
\left\{
x\in\sigma(u)
\,\middle|\,
\chi_{\mathrm{bed}}(x;\eta)
\chi_{\mathrm{slice}}(x;\eta)
\chi_{\mathrm{ik}}(x;\eta)
\chi_{\mathrm{joint}}(x;\eta)=1
\right\}.
\label{eq:feasible_set}
\end{equation}

When the profile argument is omitted, $\mathcal{X}_{\mathrm{f}}(u)$ denotes $\mathcal{X}_{\mathrm{f}}(u;\eta_{\mathrm{F}})$. Let $\omega(u)$ denote the interpreted objective. In a two-stage workflow, $\eta_{\mathrm{pat}}$ denotes the profile used for infill-pattern screening and $\eta_{\mathrm{plc}}$ denotes the profile used for orientation--placement evaluation; for a single-profile workflow, $\eta_{\mathrm{pat}}=\eta_{\mathrm{plc}}=\eta$. Each objective is ranked by the ordered priority vector
\begin{equation}
\mathbf{r}_{u}
\left(
x;\eta_{\mathrm{pat}},\eta_{\mathrm{plc}}
\right)
=
\begin{cases}
\left[
t_{\mathrm{f}}(p,\theta_{\mathrm{a}},\mathbf{c}_{\mathrm{a}};\eta_{\mathrm{pat}}),\,
j_{6}^{\max}(x;\eta_{\mathrm{plc}}),\,
\bar{j}_{6}(x;\eta_{\mathrm{plc}})
\right]^{\top},
& \text{time-minimization request},\\[1mm]
\left[
\ell_{\mathrm{e}}(p,\theta_{\mathrm{a}},\mathbf{c}_{\mathrm{a}};\eta_{\mathrm{pat}}),\,
j_{6}^{\max}(x;\eta_{\mathrm{plc}}),\,
\bar{j}_{6}(x;\eta_{\mathrm{plc}})
\right]^{\top},
& \text{material-saving request},\\[1mm]
\left[
j_{6}^{\max}(x;\eta_{\mathrm{plc}}),\,
\bar{j}_{6}(x;\eta_{\mathrm{plc}})
\right]^{\top},
& \text{motion-quality request},
\end{cases}
\qquad
x=(p,\theta,\mathbf{c}).
\label{eq:ranking_vector}
\end{equation}

For a time- or material-oriented request, the first entry of $\mathbf{r}_{u}$ screens the slicer-process candidates at the common anchor pose $(\theta_{\mathrm{a}},\mathbf{c}_{\mathrm{a}})$:
\begin{equation}
\hat p
\in
\begin{cases}
\displaystyle
\operatorname*{arg\,min}_{p:\,(p,\theta_{\mathrm{a}},\mathbf{c}_{\mathrm{a}})\in\mathcal{X}_{\mathrm{f}}(u;\eta_{\mathrm{pat}})}
t_{\mathrm{f}}(p,\theta_{\mathrm{a}},\mathbf{c}_{\mathrm{a}};\eta_{\mathrm{pat}}),
& \text{time-minimization request},\\[3mm]
\displaystyle
\operatorname*{arg\,min}_{p:\,(p,\theta_{\mathrm{a}},\mathbf{c}_{\mathrm{a}})\in\mathcal{X}_{\mathrm{f}}(u;\eta_{\mathrm{pat}})}
\ell_{\mathrm{e}}(p,\theta_{\mathrm{a}},\mathbf{c}_{\mathrm{a}};\eta_{\mathrm{pat}}),
& \text{material-saving request}.
\end{cases}
\label{eq:pattern_selection}
\end{equation}
The comparison set for the subsequent jerk-based ranking is
\begin{equation}
\mathcal{B}_{u}(\eta_{\mathrm{plc}})
=
\begin{cases}
\left\{
(\hat p,\theta,\mathbf{c})
\in
\mathcal{X}_{\mathrm{f}}(u;\eta_{\mathrm{plc}})
\right\},
& \text{time- or material-oriented request},\\[3mm]
\mathcal{X}_{\mathrm{f}}(u;\eta_{\mathrm{plc}}),
& \text{motion-quality request}.
\end{cases}
\label{eq:primary_comparison_set}
\end{equation}

The remaining entries of $\mathbf{r}_u$, namely $j_{6}^{\max}$ and then $\bar{j}_{6}$, rank candidates within $\mathcal{B}_{u}$ by a tolerance-aware lexicographic rule. For any nonempty comparison set $\mathcal{B}$, candidates whose maximum Joint-6 jerk lies within $1\%$ of the minimum form the near-tie set
\begin{equation}
\mathcal{N}_{\epsilon}(\mathcal{B};\eta)
=
\left\{
x\in\mathcal{B}
\;\middle|\;
j_{6}^{\max}(x;\eta)
\leq
(1+\epsilon)
\min_{x'\in\mathcal{B}}j_{6}^{\max}(x';\eta)
\right\},
\qquad
\epsilon=0.01.
\label{eq:j6_ranking_tolerance}
\end{equation}
The selected candidate minimizes mean absolute Joint-6 jerk within this near-tie set:
\begin{equation}
\operatorname{Sel}_{\epsilon}(\mathcal{B};\eta)
=
\operatorname*{arg\,min}_{x\in\mathcal{N}_{\epsilon}(\mathcal{B};\eta)}
\bar{j}_{6}(x;\eta),
\qquad
x^{\star}(\eta_{\mathrm{plc}})
\in
\operatorname{Sel}_{\epsilon}
\left(
\mathcal{B}_{u}(\eta_{\mathrm{plc}});
\eta_{\mathrm{plc}}
\right).
\label{eq:j6_ranking_selection}
\end{equation}

Equations~\eqref{eq:ranking_vector}--\eqref{eq:j6_ranking_selection} define the ranking rule used in the reported tasks. For time- and material-oriented requests, the leading non-jerk criterion selects the infill pattern at the anchor pose, and the Joint-6 jerk criteria select the orientation--placement candidate within $\mathcal{B}_{u}$. For a motion-quality request, the entire feasible set is ranked directly by the Joint-6 jerk criteria. For a workflow evaluated entirely with the final print profile, $\eta_{\mathrm{pat}}=\eta_{\mathrm{plc}}=\eta_{\mathrm F}$ and the reported winner is $x^{\star}=x^{\star}(\eta_{\mathrm F})$. In a profile-tagged two-stage workflow, the stage-two winner is subsequently re-evaluated under $\eta_{\mathrm F}$ for execution feasibility and final evidence; this verification does not establish that the reduced-profile ranking is preserved under $\eta_{\mathrm F}$.

\subsection{Agent protocol and workflow scoping}
\label{sec:agent_protocol}

\begin{figure}
    \centering
    \includegraphics[width=1\linewidth]{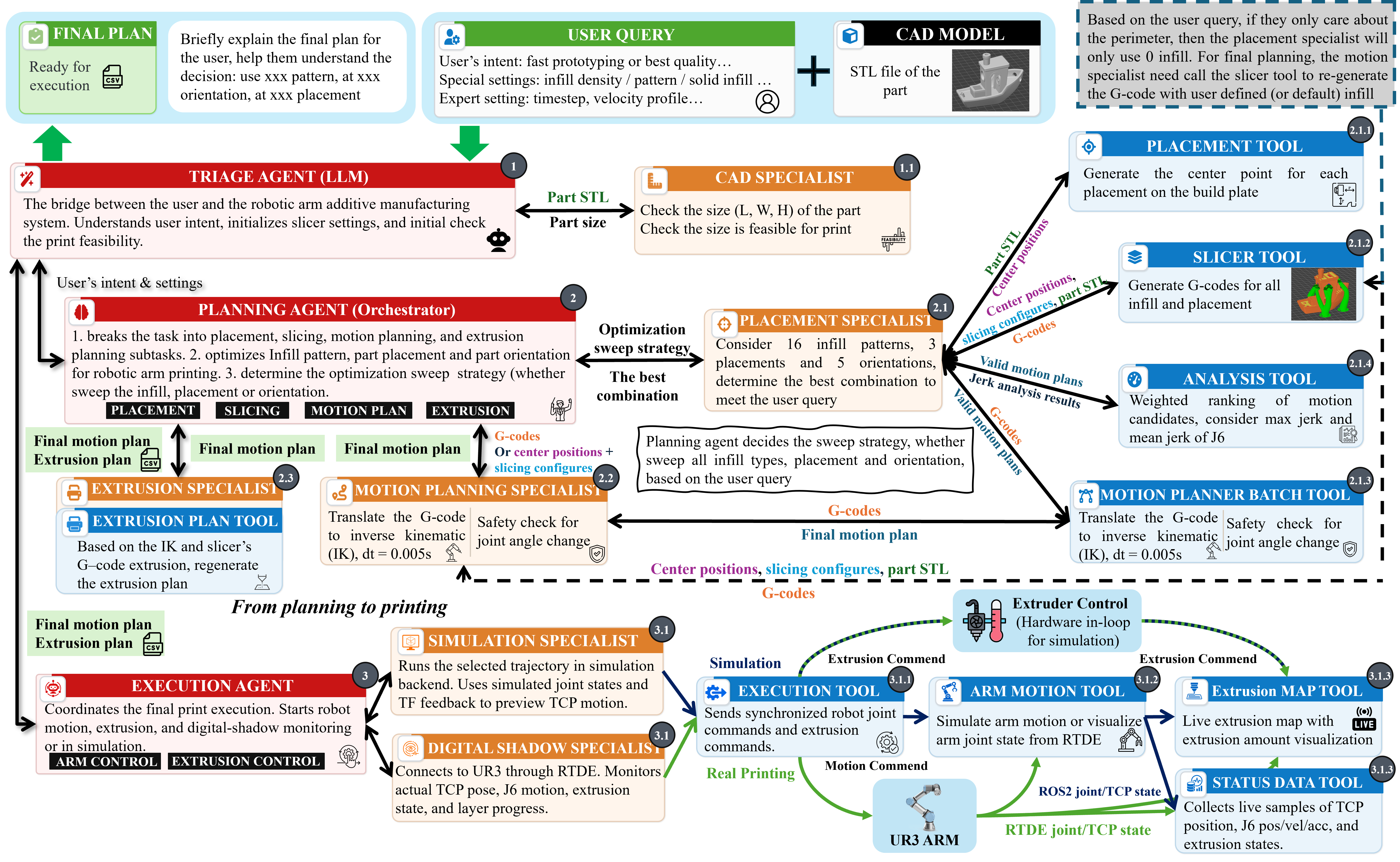}
    \caption{A-RAM agent--specialist--tool architecture for robotic AM. The LLM-based Triage Agent interprets the manufacturing objective and constraints, identifies the planning variables that require exploration, and produces a schema-constrained planning request. The Planning Agent deterministically instantiates this request as the search and evaluation workflow, specialists coordinate tools for candidate generation and verification, and the Execution Agent dispatches the selected plan to simulation or monitored physical printing.}
    \label{fig:A-RAM_robotic_arm_fff}
\end{figure}

The agent--specialist--tool architecture is shown in Fig.~\ref{fig:A-RAM_robotic_arm_fff}. The LLM-based Triage Agent interprets the free-form manufacturing request $u$, reasons over its objective and constraints, identifies the planning variables that are prescribed or require exploration, and encodes this planning intent as the structured request
\begin{equation}
\rho(u)
=
\left(
\omega,\,
\boldsymbol{\kappa},\,
\mathcal{V}_{\mathrm{fix}},\,
\mathcal{V}_{\mathrm{search}},\,
\mu
\right),
\label{eq:request_schema}
\end{equation}
where $\omega$ is the manufacturing objective interpreted from the user request, $\boldsymbol{\kappa}$ contains the interpreted user constraints together with validated process limits, $\mathcal{V}_{\mathrm{fix}}$ contains planning variables identified as prescribed by the user, $\mathcal{V}_{\mathrm{search}}$ contains variables identified as requiring exploration, and $\mu$ specifies the requested downstream execution mode, such as simulation, physical printing, or both. The LLM therefore reasons over the manufacturing intent and the planning degrees of freedom, but it does not estimate the execution-critical numerical quantities used for candidate feasibility or ranking. The resulting request is schema-validated before it reaches the Planning Agent. Invalid objectives, missing required fields, unsupported slicer settings, infeasible numeric ranges, or ambiguous execution modes trigger a repair prompt using the validation error as feedback. If the request remains invalid after a bounded number of repair attempts, the system returns the unresolved fields for human clarification instead of fabricating execution parameters. In the reported implementation, the Triage Agent is run with deterministic decoding, and numerical values used by downstream tools are either user-provided, system-defaulted, or tool-computed.

The Planning Agent deterministically instantiates the validated request $\rho(u)$ by mapping its objective, constraints, and fixed/search variables to a query-dependent search scope $\sigma(u)$, the corresponding evaluation profile when applicable, and the specialist--tool workflow. Thus, different user queries activate different search and evaluation workflows rather than passing through a single fixed candidate sweep. When the query leaves the slicer-process variables, orientation, and placement open, the Planning Agent activates a full search,
\begin{equation}
\sigma_{\mathrm{full}}(u)
=
\left\{
(p,\theta,\mathbf{c})
\,\middle|\,
p\in\mathcal{P}_{u},\;
\theta\in\Theta_{u},\;
\mathbf{c}\in\mathcal{C}_{u}(\theta)
\right\},
\qquad
\left|\sigma_{\mathrm{full}}(u)\right|
=
|\mathcal{P}_{u}|
\sum_{\theta\in\Theta_{u}}
|\mathcal{C}_{u}(\theta)|
\label{eq:full_sweep}
\end{equation}

Only in the special case $|\mathcal{C}_{u}(\theta)|=C$ for every $\theta\in\Theta_u$ does this reduce to $|\mathcal{P}_{u}|\,|\Theta_{u}|\,C$. For the default experimental grid with 16 infill-pattern candidates, five feasible orientations, and three placement centers at every orientation, Eq.~\eqref{eq:full_sweep} gives $16(5\times3)=240$ evaluated candidate jobs.

For time- or material-oriented requests, the Planning Agent can instead activate a two-stage workflow that screens infill patterns before evaluating orientation and placement. Pattern candidates are therefore first compared at a common anchor pose $(\theta_{\mathrm{a}},\mathbf{c}_{\mathrm{a}})$, with $\theta_{\mathrm{a}}=0^{\circ}$ and $\mathbf{c}_{\mathrm{a}}=\mathbf{c}_{2}$ in the reported experiments, and the selected pattern $\hat{p}$ is then passed to the orientation--placement search. This allows extrusion path length or motion-plan completion time to be evaluated once per pattern during the pattern-screening stage, while robot-specific feasibility and Joint-6 motion quality are subsequently evaluated over the orientation--placement candidates. The two workflow stages are represented as sets of profile-tagged evaluations,
\begin{align}
\mathcal{Z}_{\mathrm{pat}}(u)
&=
\left\{
(p,\theta_{\mathrm{a}},\mathbf{c}_{\mathrm{a}},\eta_{\mathrm{pat}})
\,\middle|\,
p\in\mathcal{P}_{u}
\right\},
\label{eq:pattern_screen_set}\\
\mathcal{Z}_{\mathrm{plc}}(u)
&=
\left\{
(\hat{p},\theta,\mathbf{c},\eta_{\mathrm{plc}})
\,\middle|\,
\theta\in\Theta_{u},\;
\mathbf{c}\in\mathcal{C}_{u}(\theta)
\right\},
\label{eq:placement_screen_set}\\
\mathcal{Z}_{\mathrm{screen}}(u)
&=
\mathcal{Z}_{\mathrm{pat}}(u)
\cup
\mathcal{Z}_{\mathrm{plc}}(u).
\label{eq:screen_sweep}
\end{align}

Assuming $\theta_{\mathrm{a}}\in\Theta_u$, $\mathbf{c}_{\mathrm{a}}\in\mathcal{C}_{u}(\theta_{\mathrm{a}})$, and $\hat p\in\mathcal P_u$, the general set cardinality is
\begin{equation}
\left|\mathcal{Z}_{\mathrm{screen}}(u)\right|
=
|\mathcal{P}_{u}|
+
\sum_{\theta\in\Theta_{u}}|\mathcal{C}_{u}(\theta)|
-
\mathbf{1}\!\left\{\eta_{\mathrm{pat}}=\eta_{\mathrm{plc}}\right\},
\label{eq:screen_count_general}
\end{equation}
where the indicator subtracts the single duplicated anchor evaluation only when both stages use the same evaluation profile. In the reported reduced-profile workflow, pattern screening uses the target-density profile $\eta_{\mathrm{pat}}$ and placement screening uses the zero-infill profile $\eta_{\mathrm{plc}}$, so $\eta_{\mathrm{pat}}\neq\eta_{\mathrm{plc}}$ and the two anchor evaluations are formally distinct. With 16 patterns, five orientations, and three placements per orientation, the evaluation count is therefore $16+5\times3=31$.

When the user already fixes the slicer-process candidate, the Planning Agent instead activates a fixed-process orientation--placement workflow,
\begin{equation}
\sigma_{\mathrm{place}}(u)
=
\left\{
(p_{0},\theta,\mathbf{c})
\,\middle|\,
\theta\in\Theta_{u},\;
\mathbf{c}\in\mathcal{C}_{u}(\theta)
\right\},
\qquad
|\sigma_{\mathrm{place}}(u)|
=
\sum_{\theta\in\Theta_{u}}|\mathcal{C}_{u}(\theta)|.
\label{eq:placement_only_sweep}
\end{equation}

Under the default grid, this gives $|\sigma_{\mathrm{place}}|=5\times3=15$. The full, two-stage, and fixed-process orientation--placement formulations therefore represent alternative query-conditioned search scopes selected by the Planning Agent according to the objective and the variables fixed or left open by the user.

\begin{figure}
    \centering
    \includegraphics[width=1\linewidth]{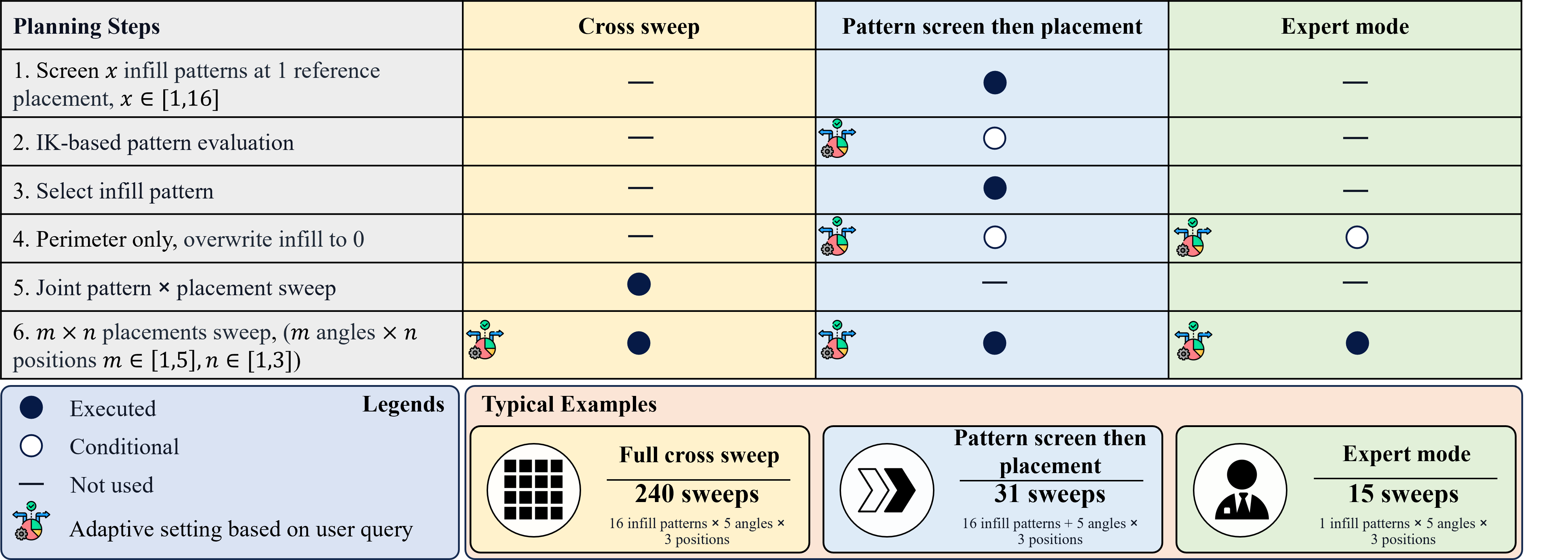}
    \caption{Representative query-conditioned sweep scopes. The Planning Agent changes $\sigma(u)$ according to query specificity, ranging from a full slicer--orientation--placement search to profile-tagged two-stage screening or expert-constrained placement evaluation.}
    \label{fig:sweep_strategy}
\end{figure}

\subsection{Candidate generation}
\label{sec:candidate_generation}

Candidate generation begins with geometric screening of $\mathcal{M}$. The CAD analysis stage extracts mesh bounds, dimensions, watertightness, volume, surface area, and the projected footprint. For each $\theta\in\Theta_{u}$, the oriented footprint is checked against the usable build-plate region with a prescribed safety margin. Feasible orientations define admissible placement-center intervals, from which the placement stage samples deterministic centers along the larger available workspace direction. This produces the orientation-dependent placement set $\mathcal{C}_{u}(\theta)$ without treating elementary bounding-box arithmetic as part of the formal method.

For each evaluated job $(x,\eta)$, with $x=(p,\theta,\mathbf{c})$, the slicer stage applies the orientation and placement to the mesh, applies the process settings associated with $p$, applies only the explicit overrides recorded in $\eta$, and exports G-code $g(x;\eta)$. The active slicer variables include infill pattern, infill density, infill angle, and solid-layer settings when they are included in $\mathcal{V}_{\mathrm{search}}$ or specified in $\mathcal{V}_{\mathrm{fix}}$. The generated G-code is parsed into deposition and travel segments, feature labels, segment-level extrusion targets, and path statistics. These artifacts provide process-side evidence, such as $\ell_{\mathrm{e}}(x;\eta)$, and the Cartesian segment sequence used by the robot-side trajectory planner.

When reduced evaluation G-code is used, $\eta$ makes the distinction between the evaluation job and the final print job explicit. For example, $\eta_{\mathrm{plc}}$ can set infill density to zero during placement screening to isolate shell motion, while $\eta_{\mathrm{F}}$ retains the final user-specified infill settings. After the reduced-profile screen selects $\theta^{\star}$ and $\mathbf{c}^{\star}$, the candidate $(\hat p,\theta^{\star},\mathbf{c}^{\star})$ is re-sliced using $\eta_{\mathrm{F}}$. The final-profile G-code is then passed through robot-motion verification and extrusion generation; reduced-profile evidence is never presented as if it were computed from the final G-code.

\subsection{Robot-side trajectory verification}
\label{sec:trajectory_verification}

The parsed G-code for evaluated job $(x,\eta)$ is represented as an ordered segment sequence $\{\mathbf{r}_{i}(\lambda;x,\eta)\}_{i=1}^{n_{s}}$, where $\lambda\in[0,1]$ parameterizes each Cartesian segment. The trajectory planner assigns boundary speeds at segment junctions, generates a constant-acceleration time parameterization, samples target TCP poses, and solves inverse kinematics sequentially. This module is used as a deterministic evaluation model for comparing candidates under identical timing assumptions; it is not claimed as a new globally optimal time-parameterization algorithm.

Let segment $i$ connect junction $i-1$ to junction $i$, with length $\ell_i$, junction speeds $v_{i-1}$ and $v_i$, acceleration bound $a_{\max}$, deceleration bound $d_{\max}$, and TCP speed bound $v_{\max}$. The feasible peak speed for the segment is
\begin{equation}
v_{i}^{\star}
=
\min
\left[
v_{\max},
\left(
\frac{
2a_{\max}d_{\max}\ell_i
+
d_{\max}v_{i-1}^{2}
+
a_{\max}v_{i}^{2}
}{
a_{\max}+d_{\max}
}
\right)^{1/2}
\right].
\label{eq:peak_speed}
\end{equation}

The segment traversal time follows from the acceleration, cruise, and deceleration phases:
\begin{equation}
t_i
=
\frac{v_{i}^{\star}-v_{i-1}}{a_{\max}}
+
\frac{v_{i}^{\star}-v_i}{d_{\max}}
+
\frac{
\left[
\ell_i
-
\frac{(v_i^{\star})^{2}-v_{i-1}^{2}}{2a_{\max}}
-
\frac{(v_i^{\star})^{2}-v_{i}^{2}}{2d_{\max}}
\right]_{+}
}{v_i^{\star}},
\qquad
t_{\mathrm{f}}(x;\eta)=\sum_{i=1}^{n_s} t_i .
\label{eq:segment_time}
\end{equation}

Here, $[\cdot]_{+}=\max(0,\cdot)$, and zero-length segments are removed before time parameterization. Equation~\eqref{eq:segment_time} covers both triangular and trapezoidal profiles through the nonnegative cruise-distance term.

To avoid stop--go motion at short G-code segments, junction speeds are refined by lookahead. Let $\mathbf{u}_{i}$ be the unit direction of segment $i$, and let $\alpha_i=\cos^{-1}(\mathbf{u}_{i}^{\top}\mathbf{u}_{i+1})$ be the turning angle at interior junction $i$. The initial junction-speed cap is $v_i^{(0)}=v_{\max}\gamma(\alpha_i)$, with $\gamma(0)=1$ and $\gamma(\pi)=0$; the implementation uses $\gamma(\alpha_i)=\max(0,\cos\alpha_i)$. With boundary speeds $v_0=v_{n_s}=0$, forward and backward feasibility passes enforce
\begin{equation}
\begin{aligned}
v_i
&\leftarrow
\min
\left(
v_i,\,
\sqrt{v_{i-1}^{2}+2a_{\max}\ell_i}
\right),
&& i=1,\ldots,n_s,\\
v_{i-1}
&\leftarrow
\min
\left(
v_{i-1},\,
\sqrt{v_i^{2}+2d_{\max}\ell_i}
\right),
&& i=n_s,\ldots,1 .
\end{aligned}
\label{eq:lookahead}
\end{equation}

After time parameterization, inverse kinematics is solved at each sampled target pose. Write the target pose as ${}^{0}\mathbf{T}_{k}=(\mathbf{R}_{k},\mathbf{p}_{k})\in SE(3)$ and the forward-kinematic pose as $FK(\mathbf q)=(\mathbf R(\mathbf q),\mathbf p(\mathbf q))$. The translational and rotational error coordinates are
\begin{equation}
\mathbf e_{p}(\mathbf q,k)
=
\mathbf p(\mathbf q)-\mathbf p_k,
\qquad
\mathbf e_{R}(\mathbf q,k)
=
\operatorname{Log}\!\left(\mathbf R_k^{\top}\mathbf R(\mathbf q)\right)^{\vee},
\label{eq:pose_error_components}
\end{equation}
and the weighted pose-error metric is
\begin{equation}
d_{SE(3)}\!\left(FK(\mathbf q),{}^{0}\mathbf T_k\right)
=
\left\|
\mathbf W_T
\begin{bmatrix}
\mathbf e_p(\mathbf q,k)\\
\mathbf e_R(\mathbf q,k)
\end{bmatrix}
\right\|_2,
\qquad
\mathbf W_T
=
\operatorname{diag}\!\left(w_p\mathbf I_3,w_R\mathbf I_3\right).
\label{eq:pose_error_metric}
\end{equation}

Let $\mathcal S_k$ be the ordered seed list supplied to the IK plugin, and let $\operatorname{IK}_{\mathrm{plug}}({}^{0}\mathbf T_k;\mathbf s)$ denote the finite set of solutions returned from seed $\mathbf s$. The enumerated, deduplicated solution set is
\begin{equation}
\mathcal Q_k^{\mathrm{enum}}(x;\eta)
=
\operatorname{Unique}_{\varepsilon_{\mathrm{dup}}}
\left(
\bigcup_{\mathbf s\in\mathcal S_k}
\operatorname{IK}_{\mathrm{plug}}
\!\left({}^{0}\mathbf T_k(x;\eta);\mathbf s\right)
\right),
\label{eq:ik_enumeration}
\end{equation}
where periodic coordinates are compared using the same wrapping convention used by the continuation stage. The accepted IK-feasible set is the subset of enumerated solutions that satisfies the external pose-error and joint-limit checks:
\begin{equation}
\mathcal{Q}_{k}^{\mathrm{ik}}(x;\eta)
=
\left\{
\mathbf{q}\in\mathcal Q_k^{\mathrm{enum}}(x;\eta)
\,\middle|\,
d_{SE(3)}
\!\left(
FK(\mathbf{q}),{}^{0}\mathbf{T}_{k}(x;\eta)
\right)
\leq
\varepsilon_{T},
\;
\mathbf{q}^{\min}\leq\mathbf{q}\leq\mathbf{q}^{\max}
\right\}.
\label{eq:ik_set}
\end{equation}

The IK routine is invoked from an ordered seed list $\mathcal S_k$. A fixed initial seed is used at the first trajectory sample, and the previously accepted configuration $\mathbf q_{k-1}$ is the leading seed at every subsequent sample. The pose-error tolerance $\varepsilon_T$, translational and rotational weights $w_p$ and $w_R$, seed-generation order, and duplicate-solution threshold $\varepsilon_{\mathrm{dup}}$ are held fixed for every candidate in a comparison set. The continuity-admissible subset is
\begin{equation}
\mathcal{Q}_{k}^{\mathrm{c}}(x;\eta)
=
\left\{
\mathbf{q}\in\mathcal{Q}_{k}^{\mathrm{ik}}(x;\eta)
\,\middle|\,
\left|
\operatorname{wrap}_{2\pi}
\!\left(
q_j-q_{k-1,j}
\right)
\right|
\leq
\delta q_{j}^{\max},
\;
j=1,\ldots,5
\right\}.
\label{eq:continuity_set}
\end{equation}

If $\mathcal{Q}_{k}^{\mathrm{ik}}(x;\eta)=\varnothing$, then $\chi_{\mathrm{ik}}(x;\eta)=0$; if $\mathcal{Q}_{k}^{\mathrm{c}}(x;\eta)=\varnothing$, then $\chi_{\mathrm{joint}}(x;\eta)=0$. Otherwise, the accepted configuration is selected by nearest-neighbor continuation:
\begin{equation}
\mathbf{q}_{k}(x;\eta)
=
\underset{\mathbf{q}\in\mathcal{Q}_{k}^{\mathrm{c}}(x;\eta)}{\operatorname{arg\,min}}
\;
\left\|
\mathbf{D}_{q}
\left(
\tilde{\mathbf{q}}-\mathbf{q}_{k-1}(x;\eta)
\right)
\right\|_{2},
\qquad
\mathbf D_q=\operatorname{diag}(d_1,\ldots,d_6).
\label{eq:ik_selection}
\end{equation}

The positive diagonal entries $d_1,\ldots,d_6$ define the fixed joint-space normalization used for nearest-neighbor continuation. The vector $\tilde{\mathbf q}$ is obtained after periodic wrist unwrapping. Specifically, Joint~6 is replaced by
\begin{equation}
\tilde{q}_{6}
=
q_6
+
2\pi
\underset{n\in\mathbb{Z}}{\operatorname{arg\,min}}
\left|
q_6+2\pi n-q_{k-1,6}
\right|.
\label{eq:j6_unwrap}
\end{equation}

The sequential inverse-kinematics stage returns, for each printing and transition segment, the accepted joint configurations $\mathbf{q}_{k}(x;\eta)$ on that segment's planner time stamps. The per-segment configurations are concatenated into a single time-ordered joint trajectory $\{t_k,\mathbf{q}_{k}(x;\eta)\}_{k=1}^{n_x}$, whose time stamps are generally nonuniform. This concatenated trajectory is then fitted, joint by joint, with a B-spline and resampled onto a candidate-specific time grid $\{t_r^{c}(x;\eta)\}_{r=0}^{N_{x,\eta}}$ with common nominal spacing $\Delta t_c=0.005\,\mathrm{s}$,
\begin{equation}
\hat{q}_{i}(t;x,\eta)
=
\sum_{\ell=1}^{n_b}
\beta_{i,\ell}(x;\eta)B_{\ell,d_{\mathrm B}}(t),
\qquad
i=1,\ldots,6,
\qquad
d_{\mathrm B}=3,
\label{eq:joint_resample_spline}
\end{equation}
where $n_b$ denotes the number of spline basis functions. Thus, the batch-ranking and final-verification stages use the same nominal resampling interval even though their original IK waypoint intervals differ. The resulting time vectors remain candidate-dependent because trajectory completion times differ across candidates. The same spline construction procedure and endpoint treatment are used throughout, with cubic degree $d_{\mathrm B}=3$. The stored trajectory that enters every downstream statistic is the spline evaluated on the corresponding candidate-specific resampling grid. This resampled joint trajectory is the object passed to $\mathcal{A}$.

The continuity gate rejects candidates with excessive wrapped joint changes in joints 1--5, while the unwrapped wrist coordinate records the orientation accommodation that remains visible in $q_6(t;x,\eta)$. This is why Joint~6 jerk is used as a sensitive metric for placement-dependent wrist motion rather than hidden inside the feasibility filter. Candidate rejection is based on deterministic candidate-level checks at irreversible stages: build-plate fit, slicer success, IK existence, joint limits, and joint-continuity limits; execution-parameter ranges are validated earlier at the request-schema stage. Here, feasibility refers to the stated sampled kinematic checks, and the timing model estimates motion-plan duration; these checks do not constitute a complete verification of robot dynamics or collision clearance.

\subsection{Motion-quality metric}
\label{sec:motion_quality_metric}

Maximum absolute Joint-6 jerk is used as the primary ranking criterion. Placement changes the IK mapping of the same prescribed TCP trajectory, so the placement-dependent motion-quality difference is evaluated in joint space. Over the current workspace, $q_5$ is nearly stationary and $q_1$--$q_4$ exhibit substantially different jerk magnitudes, so a full-joint metric would require additional normalization or weighting to avoid domination by large-magnitude joints. Joint~6 is used as a consistent local wrist-motion indicator; its jerk does not directly measure whole-arm motion quality or deposited-part quality. The maximum value is used to capture the worst local motion transient, which may be diluted by trajectory-averaged measures.

The motion-quality metric is computed from a single jerk definition used for all candidate rankings and reported tables. Let $\{t_r,q_{6}(t_r;x,\eta)\}_{r=0}^{N_{x,\eta}}$ denote the unwrapped Joint~6 coordinate of candidate $x$ on its candidate-specific resampling grid, where $t_r$ denotes the corresponding recorded time stamp from Eq.~\eqref{eq:joint_resample_spline}. All candidates use the same nominal resampling interval of $0.005\,\mathrm{s}$, although their trajectory durations and therefore their numbers of samples may differ. The sampled time vectors are nominally uniform but may contain floating-point representation differences, so all derivatives are taken with respect to the recorded time vector rather than a nominal step size.

Joint~6 jerk is obtained by applying a second-order-accurate central first-difference operator $\mathcal{D}$ three times to the sampled position. At an interior node, with $h_r^{-}=t_r-t_{r-1}$ and $h_r^{+}=t_{r+1}-t_r$,
\begin{equation}
(\mathcal{D}y)_r
=
\frac{
(h_r^{-})^{2}y_{r+1}
+
\left[(h_r^{+})^{2}-(h_r^{-})^{2}\right]y_r
-
(h_r^{+})^{2}y_{r-1}
}{
h_r^{-}h_r^{+}(h_r^{-}+h_r^{+})
},
\label{eq:central_diff_op}
\end{equation}
with second-order one-sided formulas at the endpoints, and
\begin{equation}
j_{6}(t_r;x,\eta)=\bigl(\mathcal{D}^{3}q_{6}\bigr)_r,
\label{eq:j6_jerk}
\end{equation}
i.e. the third successive numerical derivative $\dot q_{6}=\mathcal{D}q_{6}$, $\ddot q_{6}=\mathcal{D}\dot q_{6}$, $j_{6}=\mathcal{D}\ddot q_{6}$. The same differencing convention is applied to every candidate in a comparison set.

The reported maximum and mean absolute jerk are
\begin{align}
j_{6}^{\max}(x;\eta)
&=
\max_{0\le r\le N_{x,\eta}}
|j_{6}(t_r;x,\eta)|,
\label{eq:jerk_max_grid}\\
\bar{j}_{6}(x;\eta)
&=
\frac{1}{N_{x,\eta}+1}
\sum_{r=0}^{N_{x,\eta}}
|j_{6}(t_r;x,\eta)|.
\label{eq:jerk_mean_grid}
\end{align}

Because every candidate is differenced using the same nominal $0.005\,\mathrm{s}$ resampling interval, there is no coarse/fine distinction in the jerk computation itself: the batch and final stages differ only in the waypoint interval used to generate the trajectory, not in the nominal interval used for jerk evaluation. For a motion-quality request, $\mathcal{B}_{u}(\eta)=\mathcal{X}_{\mathrm{f}}(u;\eta)$ in Eq.~\eqref{eq:primary_comparison_set}; the selected candidate is therefore obtained only by the one-percent maximum-jerk near-tie rule in Eq.~\eqref{eq:j6_ranking_tolerance}, followed by minimization of mean absolute jerk in Eq.~\eqref{eq:j6_ranking_selection}. This is the sole motion-quality ranking rule.

\subsection{Extrusion synchronization}
\label{sec:extrusion_synchronization}

The slicer provides segment-level extrusion targets, whereas the robot executes a timestep-level trajectory after velocity-profile generation and IK selection. The Extrusion Specialist therefore regenerates extrusion commands on the final robot-motion grid. Let $e_s$ be the extrusion amount assigned by the G-code to deposition segment $s$, and let $\mathcal{I}_{s}(x)$ be the set of robot-trajectory timesteps assigned to that segment. For timestep $k\in\mathcal{I}_{s}(x)$, with TCP travel increment $d_k(x)$, the synchronized extrusion command is
\begin{equation}
e_k(x)
=
\frac{d_k(x)}
{\sum_{r\in\mathcal{I}_{s}(x)}d_r(x)}
\,e_s,
\qquad
k\in\mathcal{I}_{s}(x).
\label{eq:extrusion_redistribution}
\end{equation}

Thus, $\sum_{k\in\mathcal{I}_{s}(x)}e_k(x)=e_s$ by construction, while travel moves receive $e_k(x)=0$ even when TCP speed is nonzero. Optional feature-level scaling is represented as $e_k(x)\leftarrow\alpha_{\ell}e_k(x)$ for samples belonging to feature class $\ell$, such as infill, perimeter, or solid layers. The final execution file stores the time vector, joint trajectory, TCP increments, deposition flag, feature label, cumulative extrusion target, and synchronized extrusion increments.

\begin{figure}
    \centering
    \includegraphics[width=1\linewidth]{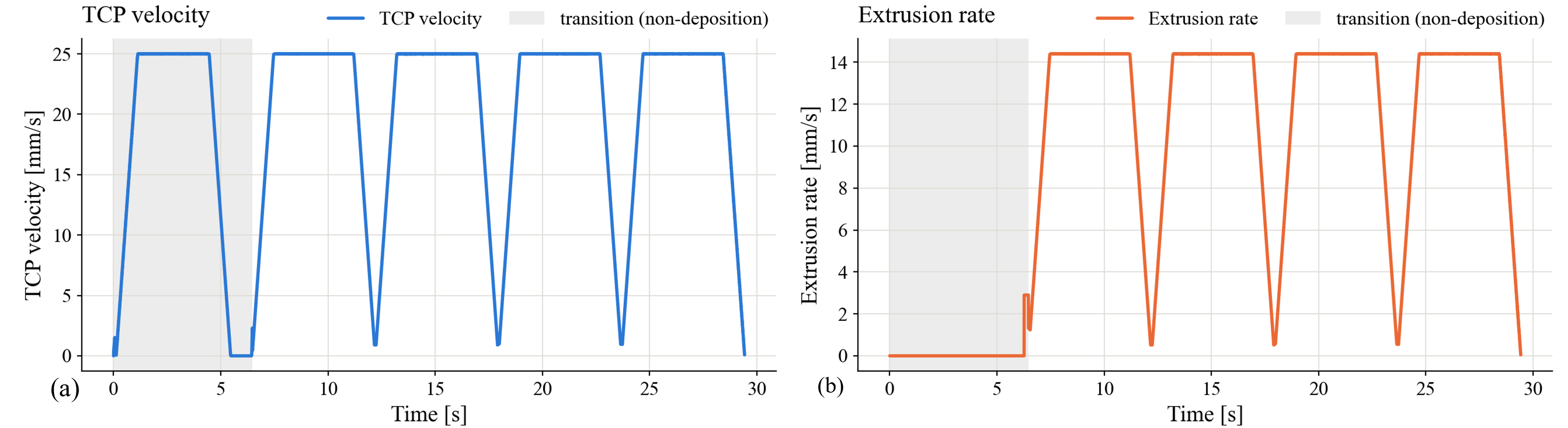}
    \caption{Motion-synchronized extrusion regeneration. Travel moves may have nonzero TCP speed with zero extrusion, whereas deposition segments redistribute the G-code extrusion target over the final robot-motion timesteps in proportion to local TCP travel distance.}
    \label{fig:extrusion_speed_relation}
\end{figure}

\subsection{Execution monitoring and traceability}
\label{sec:execution_monitoring}

The Execution Agent dispatches the selected plan either to the simulation branch or to physical printing. Both branches consume the same selected motion--extrusion trajectory. In simulation, the trajectory is executed through the ROS~2 mock-hardware branch for kinematic preview. In physical printing, the joint trajectory is sent to the UR3 controller and the synchronized extrusion targets are sent to the extrusion controller. The digital-shadow branch samples live TCP pose, Joint~6 state, extrusion command state, deposition flag, feature label, layer index, and matched trajectory index, and aligns these states with the selected plan. The role of this layer is traceability: planning tools select the candidate before execution, and the digital shadow records whether the simulated or physical execution follows the selected trajectory and extrusion command stream. Forward kinematics is used as a diagnostic reconstruction layer. With the standard flange-to-tool convention, the TCP transform is
\begin{equation}
{}^{0}\mathbf{T}_{\mathrm{tcp},k}
=
{}^{0}\mathbf{T}_{6,k}
{}^{6}\mathbf{T}_{\mathrm{tcp}},
\qquad
\mathbf{r}_{\mathrm{tcp},k}
=
\mathbf{r}_{6,k}
+
\mathbf{R}_{6,k}\,{}^{6}\mathbf{r}_{\mathrm{tcp}},
\label{eq:tcp_transform}
\end{equation}
where ${}^{6}\mathbf{r}_{\mathrm{tcp}}$ is defined from the Joint-6 flange frame to the nozzle TCP. This convention fixes the sign of the tool-offset term and keeps the forward-kinematic reconstruction consistent with the flange-to-tool transform.

The monitored extrusion state in the current implementation is command-side rather than encoder-measured filament displacement. Consequently, layer-wise deposition visualization is interpreted as a nominal command-side proxy, not as measured bead geometry. For two consecutive samples with nonzero TCP travel, the displayed bead-width proxy is computed from volume conservation as
\begin{equation}
w^{\mathrm{cmd}}_{b,k}
=
\frac{e_k A_f}{h\,\Delta s_k},
\qquad
A_f
=
\frac{\pi d_f^{2}}{4},
\label{eq:bead_width_proxy}
\end{equation}
where $e_k$ is the commanded filament-feed increment, $A_f$ is the filament cross-sectional area, $d_f$ is the filament diameter, $h$ is the layer height, and $\Delta s_k$ is the TCP travel between samples. This signal supports visualization and post-run comparison against the selected plan, but it is not used as a closed-loop bead-width measurement.

\subsection{Implementation on the robotic-arm FFF cell}
\label{sec:implementation_cell}

The implemented system uses a six-axis UR3 robotic arm equipped with a fused-filament extrusion end effector. PrusaSlicer CLI generates candidate G-code, ROS~2 and MoveIt support execution and visualization, and RTDE provides live robot-state access during physical printing. Batch candidate ranking uses $\Delta t_{\mathrm{batch}}=0.01\,\mathrm{s}$ for efficient screening, and the selected job is regenerated at $\Delta t_{\mathrm{final}}=0.005\,\mathrm{s}$; both are resampled using the same nominal $0.005\,\mathrm{s}$ interval before jerk evaluation, as defined in Eq.~\eqref{eq:joint_resample_spline}. The default build region is $160\,\mathrm{mm}\times160\,\mathrm{mm}$ with a $2\,\mathrm{mm}$ safety margin, and the default orientation set is $\Theta=\{0^{\circ},30^{\circ},45^{\circ},60^{\circ},90^{\circ}\}$. The numerical values of $v_{\max}$, $a_{\max}$, $d_{\max}$, $\delta q_{j}^{\max}$, layer height, and extrusion width are fixed by the experimental configuration and reported with the case-study settings.

The IK acceptance settings, seed-ordering rule, joint-space scaling, spline construction procedure, and nominal jerk-evaluation interval are held fixed across all candidates evaluated within the same comparison. The LLM used in the reported implementation is Qwen3-8B without task-specific fine-tuning. Because execution-critical numerical evaluation remains tool-based, the language model can be replaced or upgraded without changing the formal planning problem or quantitative verification pipeline.

\section{Experimental Design and Results}
\label{sec:results}

This section evaluates A-RAM through three case studies comprising nine query-conditioned planning tasks. Across all experiments, the layer height is fixed at $1\,\mathrm{mm}$, the extrusion width is fixed at $2.5\,\mathrm{mm}$, the maximum TCP speed is set to $v_{\max}=25\,\mathrm{mm\,s^{-1}}$, and the acceleration and deceleration limits are set to $a_{\max}=d_{\max}=25\,\mathrm{mm\,s^{-2}}$. The maximum allowable joint change between consecutive IK samples is set to $\delta q_{j}^{\max}=0.35\,\mathrm{rad}$ for joints $j=1,\ldots,5$. Joint-6 jerk is computed from the unwrapped wrist trajectory defined in Section~\ref{sec:motion_quality_metric}; $j_{6}^{\max}$ and $\bar{j}_{6}$ denote the maximum and mean absolute Joint-6 jerk, respectively, and are reported in $\mathrm{rad\,s^{-3}}$. 

Case Study~I uses a Dogbone Type~IV specimen to compare an expert-specified query with a goal-only query that asks to minimize Joint-6 jerk. Case Study~II uses a five-layer Husky Head to compare material-saving and time-minimization objectives under 90\% and 100\% infill densities. Case Study~III compares three geometries under the same perimeter-focused motion-quality query to test whether an orientation--placement rule transfers across parts. Table~\ref{tab:results_overview} summarizes the nine planning tasks and their search scopes.

\begin{table}[t]
\centering
\caption{Summary of the query-conditioned planning tasks evaluated in this section.}
\label{tab:results_overview}
\small
\setlength{\tabcolsep}{4pt}
\begin{tabular}{l l l c l}
\hline
Case & Part & Objective & Candidate scope & Execution mode \\
\hline
I-a & Dogbone Type~IV & Expert-specified plan & $1\times1\times3$ & Planning output \& digital shadow \\
I-b & Dogbone Type~IV & Minimize $j_{6}^{\max}$ & $16\times5\times3$ & Planning output \\
II-a & Husky Head, 5-layer & Save filament, 90\% infill & $16+5\times3$ & Planning output \& digital shadow\\
II-b & Husky Head, 5-layer & Fastest, 90\% infill & $16+5\times3$ & Planning output \& digital shadow \\
II-c & Husky Head, 5-layer & Save filament, 100\% infill & $16+5\times3$ & Planning output \& digital shadow \\
II-d & Husky Head, 5-layer & Fastest, 100\% infill & $16+5\times3$ & Planning output \& digital shadow \\
III-a & Gator Head, 4-layer & Minimize $j_{6}^{\max}$ & $1\times5\times3$ & Planning output \& physical print \\
III-b & Husky Head, 4-layer & Minimize $j_{6}^{\max}$ & $1\times5\times3$ & Planning output \& physical print \\
III-c & Husky Head, 5-layer & Minimize $j_{6}^{\max}$ & $1\times5\times3$ & Planning output \\
\hline
\end{tabular}
\end{table}

\subsection{Case Study I: Dogbone Type IV}
\label{sec:case1_dogbone}

\paragraph{Expert-specified query.}
This task evaluates an expert-specified request in which the user prescribes the main process parameters: \textit{gyroid} infill, 50\% infill density, 25\% infill overlap, a $45^\circ$ infill angle, a $45^\circ$ part orientation, one top solid layer, one bottom solid layer, and physical printing after planning. Because the infill pattern and part orientation are fixed, A-RAM does not perform a global pattern--orientation sweep. The Planning Agent treats the request as an expert-constrained workflow and evaluates only the three candidate placements available at the prescribed $45^\circ$ orientation. The selected plan preserves all user-defined slicing settings and chooses placement C3. A final single-candidate rerun with $\Delta t_{\mathrm{final}}=0.005\,\mathrm{s}$ then generates the execution-ready motion and extrusion plans.

\begin{tcolorbox}[
    enhanced,
    breakable,
    colback=gray!4,
    colframe=black!70,
    title={Agentic execution trace for the expert-specified Dogbone query},
    fonttitle=\bfseries,
    sharp corners,
    boxrule=0.6pt,
    left=2mm,right=2mm,top=1mm,bottom=1mm
]
\small
\textbf{User query:} Gyroid infill, 50\% density, 25\% overlap, $45^\circ$ fill angle, $45^\circ$ part orientation, one top layer, one bottom layer, and print the part after planning.\\
\textbf{Interpreted objective:} \texttt{custom}.\\
\textbf{Selected workflow:} \texttt{fixed\_pattern\_then\_placement}.\\
\textbf{Search space:} 1 infill pattern $\times$ 1 orientation $\times$ 3 placements = 3 candidates.\\
\textbf{Valid candidates:} 3 motion-planning cases satisfy the current IK and continuity constraints.\\
\textbf{Selected plan:} \textit{gyroid} infill, 50\% density, $45^\circ$ part orientation, placement C3.\\
\textbf{Final rerun:} single-candidate rerun with $\Delta t_{\mathrm{final}}=0.005\,\mathrm{s}$.\\
\textbf{Execution decision:} execution-ready motion and extrusion plans generated.
\end{tcolorbox}

\begin{figure}
    \centering
    \includegraphics[width=\textwidth]{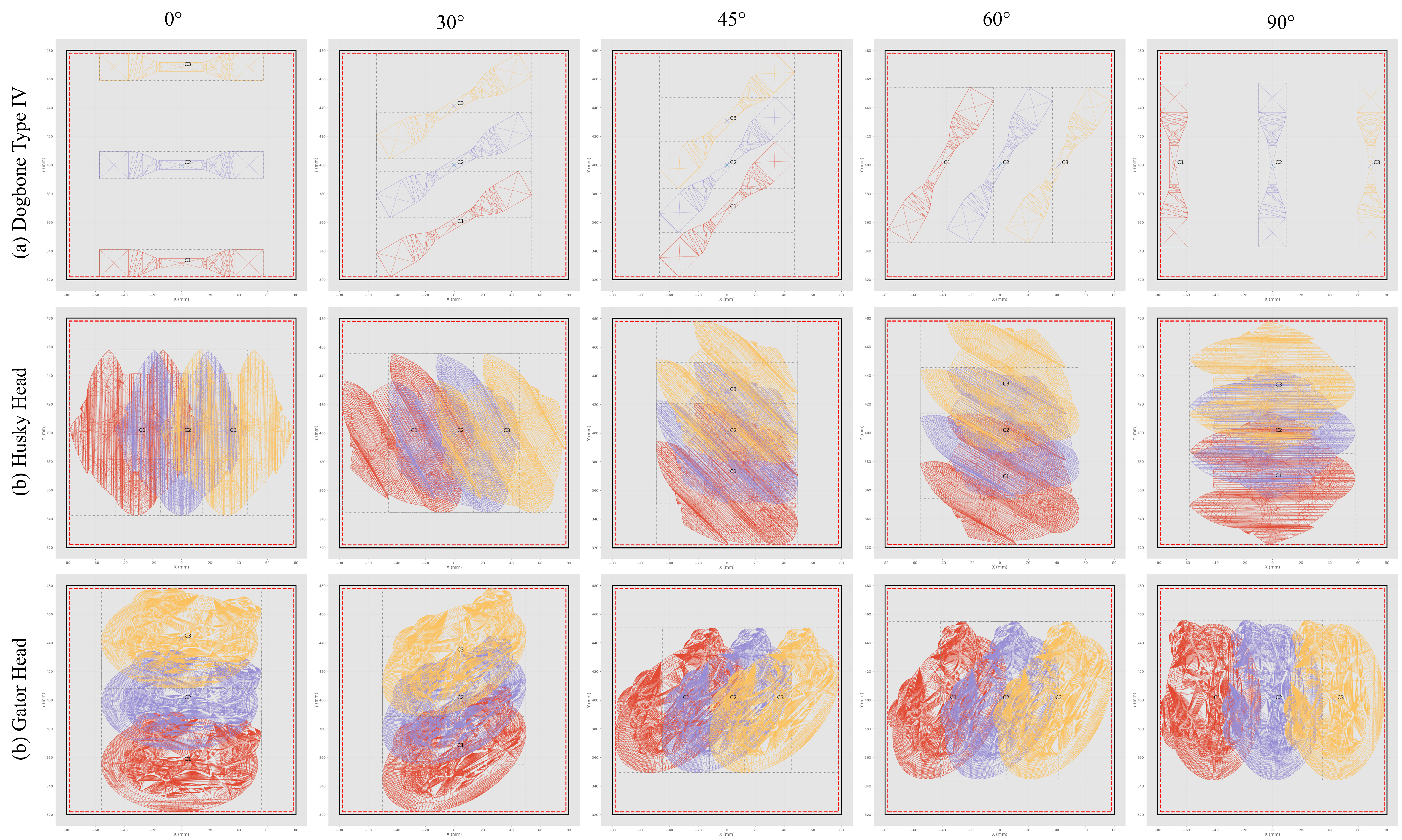}
    \caption{Combined candidate-generation figure. Orientation-dependent placement candidates for (a) Dogbone Type IV, (b) Husky Head, and (c) Gator Head. Each column corresponds to one tested part orientation, and the colored overlays denote the three generated placement centers, C1--C3, within the build region. The dashed red rectangle indicates the allowable placement region used during candidate generation.}
    \label{fig:combined_candidate_generation}
\end{figure}

With the part orientation prescribed as $45^\circ$, the Placement Tool restricts candidate generation to three placement centers, C1--C3, as shown in Fig.~\ref{fig:combined_candidate_generation} (a, $45^\circ$). The rotated footprint satisfies the build-region constraint, so all three placements are passed to slicing and robot-side evaluation.

\begin{figure}
    \centering
    \includegraphics[width=0.5\linewidth]{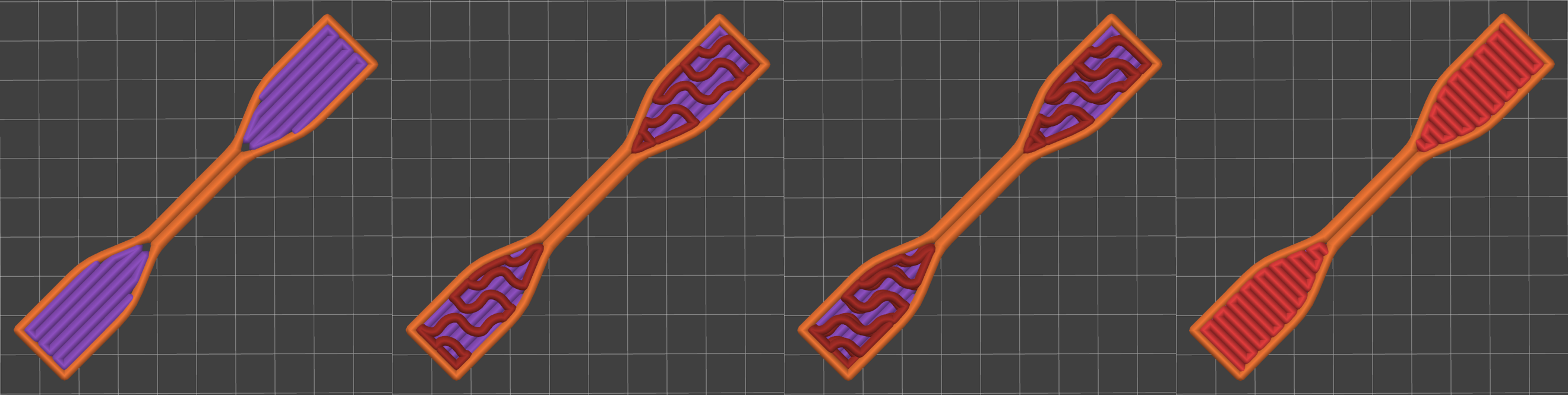}
    \caption{Representative slicer-generated layer patterns for the expert-specified Dogbone query. The orange paths denote perimeters, the red paths denote gyroid infill, and the purple paths denote top or bottom solid layers generated with the prescribed $45^\circ$ fill angle.}
    \label{fig:dogbone_expert_layers}
\end{figure}

The Slicer Tool converts each placement into G-code using the prescribed \textit{gyroid} infill, 50\% density, 25\% overlap, $45^\circ$ infill angle, one top solid layer, and one bottom solid layer. Figure~\ref{fig:dogbone_expert_layers} shows representative layers extracted from the generated G-code. These G-code paths are the direct inputs to robot-side motion planning and jerk evaluation.

Table~\ref{tab:dogbone_expert_ranking} reports the placement ranking. Because the infill pattern, slicer settings, and part orientation are fixed, the changes in Joint-6 jerk are caused by workspace placement. Placement C3 minimizes $j_{6}^{\max}$ among the three feasible candidates, reducing it from $606.7101\,\mathrm{rad\,s^{-3}}$ for C1 to $543.6724\,\mathrm{rad\,s^{-3}}$, a 10.4\% reduction. The same placement reduces $\bar{j}_{6}$ from $4.4117\,\mathrm{rad\,s^{-3}}$ to $3.6033\,\mathrm{rad\,s^{-3}}$, an 18.3\% reduction.

\begin{table}[t]
\centering
\caption{Joint-6 jerk ranking for the expert-specified Dogbone query.}
\label{tab:dogbone_expert_ranking}
\small
\begin{tabular}{c c c c c c}
\hline
Rank & Orientation & Placement & Infill pattern & $j_{6}^{\max}$ $(\mathrm{rad\,s^{-3}})$ & $\bar{j}_{6}$ $(\mathrm{rad\,s^{-3}})$\\
\hline
1 & $45^\circ$ & C3 & gyroid & 543.6724 & 3.6033 \\
2 & $45^\circ$ & C2 & gyroid & 576.1122 & 3.9878 \\
3 & $45^\circ$ & C1 & gyroid & 606.7101 & 4.4117 \\
\hline
\end{tabular}
\end{table}

\begin{figure}
    \centering
    \includegraphics[width=\textwidth]{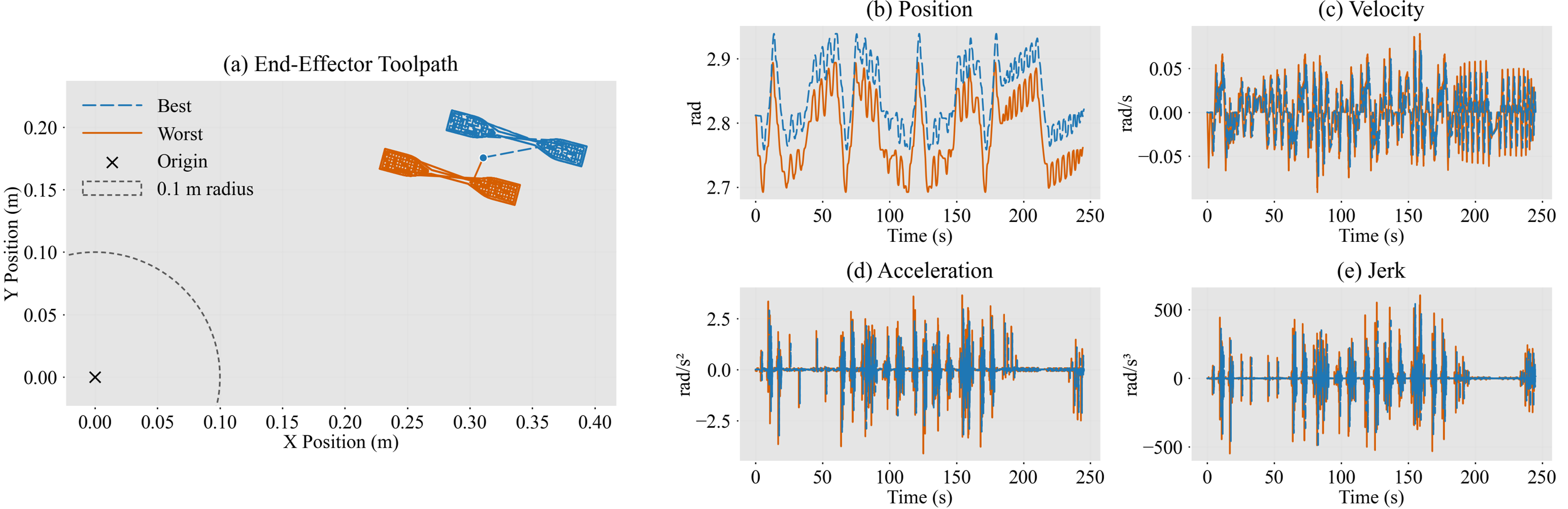}
    \caption{Comparison of the best and worst placements for the expert-specified Dogbone query. The best candidate is \textit{gyroid}, $45^\circ$, C3, while the worst candidate is \textit{gyroid}, $45^\circ$, C1. (a) End-effector XY toolpaths, including the robot origin and the reference arc with a radius of $0.1$~m. (b)--(e) Joint-6 position, velocity, acceleration, and jerk profiles, respectively.}
    \label{fig:dogbone_expert_best_worst_compare}
\end{figure}

The spatial and joint-motion comparisons in Fig.~\ref{fig:dogbone_expert_best_worst_compare} further contextualize the ranking. As shown in Fig.~\ref{fig:dogbone_expert_best_worst_compare} (a), the best and worst candidates use the same G-code pattern and part orientation, so their local deposition geometries are identical. However, the toolpaths are located at different positions within the robot workspace, leading to different induced joint trajectories. The resulting Joint-6 position, velocity, acceleration, and jerk responses are shown in Fig.~\ref{fig:dogbone_expert_best_worst_compare} (b)--(e), with the best placement exhibiting lower Joint-6 jerk than the worst placement.

After C3 is selected, the Motion Planning Specialist performs the final rerun and the Extrusion Specialist regenerates the synchronized extrusion plan. 

\paragraph{Goal-only query.}
The second Dogbone task evaluates a goal-only request in which the user specifies only the objective of minimizing Joint-6 jerk. The LLM-based Triage Agent interprets the request as \texttt{min\_j6\_jerk}. Since no infill pattern, part orientation, or placement is fixed by the user, the Planning Agent selects the full cross sweep. The resulting search evaluates 16 infill patterns, five orientations, and three placements, giving 240 candidate combinations. Under the current robotic-arm constraints, 239 candidates are valid and one candidate has no feasible IK solution. The selected plan uses \textit{stars} infill, a $30^\circ$ part orientation, and placement C3.

\begin{tcolorbox}[
    enhanced,
    breakable,
    colback=gray!4,
    colframe=black!70,
    title={Agentic execution trace for the goal-only Dogbone query},
    fonttitle=\bfseries,
    sharp corners,
    boxrule=0.6pt,
    left=2mm,right=2mm,top=1mm,bottom=1mm
]
\small
\textbf{User query:} I want to minimize Joint-6 jerk.\\
\textbf{Interpreted objective:} \texttt{min\_j6\_jerk}.\\
\textbf{Selected workflow:} \texttt{full\_cross\_sweep}.\\
\textbf{Search space:} 16 infill patterns $\times$ 5 orientations $\times$ 3 placements = 240 candidates.\\
\textbf{Valid candidates:} 239 motion-planning cases satisfy the current constraints.\\
\textbf{Invalid candidates:} 1 candidate has no feasible IK solution.\\
\textbf{Selected plan:} \textit{stars} infill, $30^\circ$ part orientation, placement C3.\\
\textbf{Final rerun:} single-candidate rerun with $\Delta t_{\mathrm{final}}=0.005\,\mathrm{s}$.\\
\textbf{Planning output:} execution-ready motion and extrusion plans generated.
\end{tcolorbox}

Because the user leaves orientation and placement open, the Placement Tool enumerates candidates across the tested orientations, as shown in Fig.~\ref{fig:combined_candidate_generation} (a). The Slicer Tool then generates G-code for the 16 infill patterns considered in the sweep. 

Table~\ref{tab:dogbone_goalonly_ranking} reports representative candidates from the Joint-6 jerk ranking of the 239 valid plans. The selected plan, \textit{stars}, $30^\circ$, C3, reduces $j_{6}^{\max}$ from $760.800\,\mathrm{rad\,s^{-3}}$ for the least favorable valid candidate to $353.593\,\mathrm{rad\,s^{-3}}$, corresponding to a 53.5\% reduction. It also reduces $\bar{j}_{6}$ from $5.524\,\mathrm{rad\,s^{-3}}$ to $2.854\,\mathrm{rad\,s^{-3}}$, corresponding to a 48.3\% reduction. The candidates are ranked using the tolerance-aware lexicographic rule defined in Eqs.~\eqref{eq:j6_ranking_tolerance}--\eqref{eq:j6_ranking_selection}. For the global ranking in Table~\ref{tab:dogbone_goalonly_ranking}, the candidate set is the full set of 239 feasible plans. Under this global ranking, no additional candidate falls within 1\% of the minimum $j_{6}^{\max}$ in this evaluation.

\begin{table}[t]
\centering
\setlength{\tabcolsep}{4pt}
\caption{Representative lexicographic Joint-6 jerk ranking for the goal-only Dogbone query.}
\label{tab:dogbone_goalonly_ranking}
\small
\begin{tabular}{c c c c c c}
\hline
Rank & Orientation & Placement & Infill pattern & $j_{6}^{\max}$ $(\mathrm{rad\,s^{-3}})$ & $\bar{j}_{6}$ $(\mathrm{rad\,s^{-3}})$\\
\hline
1   & $30^\circ$ & C3 & stars              & 353.593 & 2.854 \\
2   & $30^\circ$ & C3 & rectilinear        & 360.214 & 2.882 \\
3   & $30^\circ$ & C3 & lightning          & 361.884 & 2.632 \\
4   & $30^\circ$ & C3 & line               & 366.272 & 3.111 \\
5   & $30^\circ$ & C3 & archimedeanchords  & 375.162 & 3.032 \\
\hline
235 & $0^\circ$  & C1 & cubic              & 751.279 & 4.193 \\
236 & $0^\circ$  & C1 & grid               & 752.137 & 3.806 \\
237 & $0^\circ$  & C1 & triangles          & 752.470 & 4.079 \\
238 & $0^\circ$  & C1 & archimedeanchords  & 759.194 & 4.645 \\
239 & $0^\circ$  & C1 & honeycomb          & 760.800 & 5.524 \\
\hline
\end{tabular}
\end{table}

\begin{figure}
\centering
\includegraphics[width=1\linewidth]{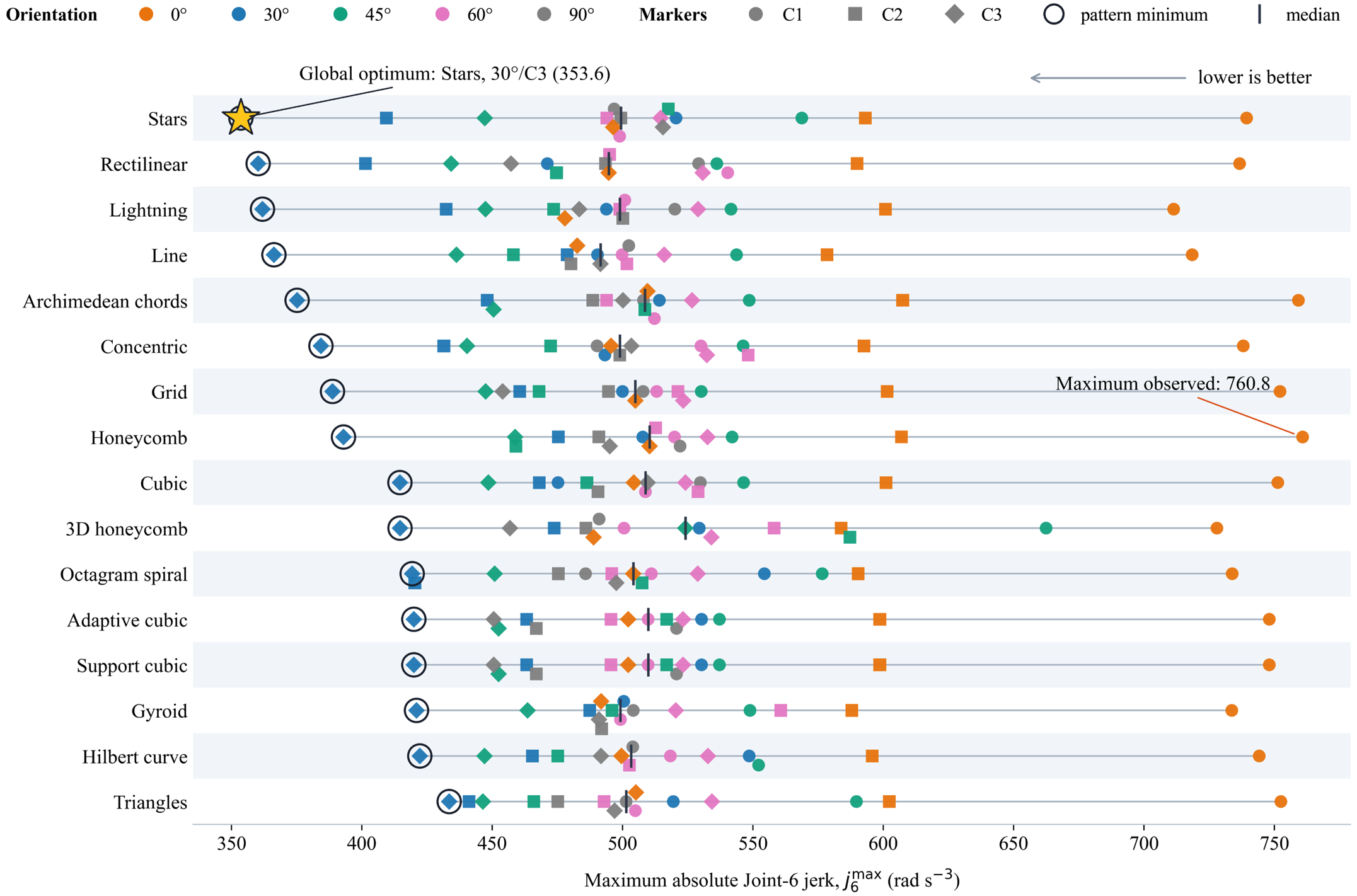}
\caption{Ranked candidate landscape for the goal-only Dogbone search. Each row shows all feasible orientation--placement candidates for one infill pattern. Horizontal lines span the within-pattern range, vertical ticks mark the medians, color denotes orientation, and marker shape denotes placement. Hollow markers identify within-pattern minima, and the gold star marks the selected candidate over the evaluated feasible set. One candidate, Hilbert curve at $90^\circ$/C2, was infeasible and is omitted.}
\label{fig:Ranked_Candidate_Landscape}
\end{figure}

The full search landscape is summarized in Fig.~\ref{fig:Ranked_Candidate_Landscape}, which includes all 239 feasible orientation--placement candidates across the 16 infill patterns. Each row shows the candidate distribution for one pattern, with marker color denoting orientation and marker shape denoting placement. The horizontal line spans the within-pattern range, the vertical tick marks the median, and the hollow marker identifies the minimum maximum absolute Joint-6 jerk within that pattern. The minimum over the evaluated feasible set is obtained for \textit{stars} at $30^\circ$/C3, with $j_{6}^{\max}=353.6\,\mathrm{rad\,s^{-3}}$. Notably, the raw maximum-jerk minimum for every infill pattern occurs at $30^\circ$/C3, indicating a consistent orientation--placement preference under this metric for the present search.

After the selected candidate is identified, the Motion Planning Specialist performs the final single-candidate rerun with $\Delta t_{\mathrm{final}}=0.005\,\mathrm{s}$. This task stops at the planning stage. The result demonstrates that a goal-only motion-quality request activates a broader candidate space than an expert-specified request and identifies the lowest-jerk feasible plan within the evaluated pattern--orientation--placement grid.

\subsection{Case Study II: Husky Head}
\label{sec:case2_husky}

This case study uses a five-layer Husky Head model to evaluate four related user queries formed by two objectives, material saving and time minimization, under two prescribed final infill densities, 90\% and 100\%. These percentages denote nominal slicer settings; material-saving comparisons use extrusion-path length without imposing equal achieved volume fraction or mechanical performance across patterns. A representative query is: "Print the Husky Head with 90\% infill while minimizing material consumption and prioritize the robot-motion quality of the perimeter and top/bottom solid-infill trajectories." The other three queries follow the same structure with the objective and infill density changed accordingly. The prescribed 90\% or 100\% infill density is retained for the final print, whereas orientation--placement selection focuses on perimeter and top/bottom solid-infill motion. These regions define the external boundary and visible solid surfaces of the part, while internal infill is enclosed within the part and primarily supports subsequent layers. At high infill densities, internal infill can constitute a large fraction of the total trajectory and could therefore dominate the placement metric, biasing selection away from the perimeter and solid-surface trajectories targeted by the query. Internal infill is consequently excluded only during orientation--placement selection and restored for the final print. The resulting orientation--placement pair is selected among the evaluated candidates using the query-defined reduced-profile motion-quality criterion.

The four compressed execution records are summarized in Table~\ref{tab:husky_trace_summary}. The same staged workflow is used in all four cases, but the pattern-screening criterion changes with the objective. The material-saving objective selects the shortest extrusion path from the G-code summary, whereas the time-minimization objective selects the shortest motion-plan end time at the common anchor pose. 

\begin{table}[t]
\centering
\caption{Agentic execution trace summary for the four Husky Head queries.}
\label{tab:husky_trace_summary}
\small
\setlength{\tabcolsep}{3pt}
\begin{tabular}{l c l l l l}
\hline
Task & Density & Interpreted objective & Pattern-screen criterion & Placement-evaluation setting & Selected final plan \\
\hline
1 & 90\% & \texttt{save\_filament} & shortest extrusion path & infill->0\%, with solid layers & \textit{lightning}, $90^\circ$, C3 \\
2 & 90\% & \texttt{fastest} & shortest motion-plan time & infill->0\%, with solid layers & \textit{supportcubic}, $90^\circ$, C3 \\
3 & 100\% & \texttt{save\_filament} & shortest extrusion path & infill->0\%, with solid layers & \textit{octagramspiral}, $90^\circ$, C3 \\
4 & 100\% & \texttt{fastest} & shortest motion-plan time & infill->0\%, with solid layers & \textit{rectilinear}, $90^\circ$, C3 \\
\hline
\end{tabular}
\end{table}

Because the user does not prescribe a fixed placement or orientation, the Placement Tool first generates the feasible orientation--placement candidates shown in Fig.~\ref{fig:combined_candidate_generation} (b). This orientation--placement set is shared by the four Husky tasks.

\begin{figure}
    \centering
    \includegraphics[width=\linewidth]{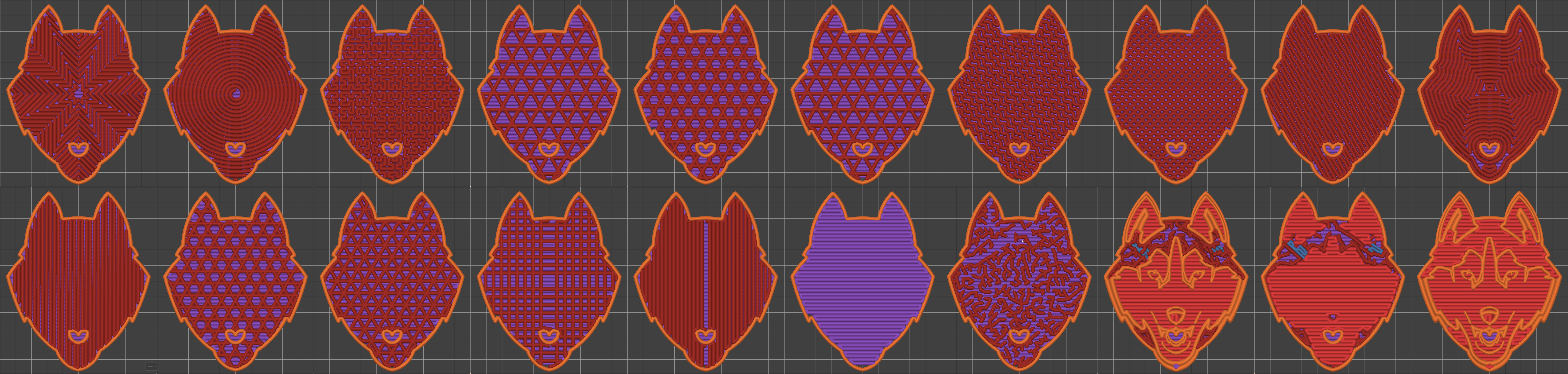}
    \caption{Slicer-generated G-code visualization under the 90\% infill-density setting. The orange paths denote perimeters, the purple and light-red paths denote bottom and top solid regions, respectively, and the dark-red paths denote infill. In the first row, from left to right, the infill patterns are \textit{octagramspiral}, \textit{archimedeanchords}, \textit{hilbertcurve}, \textit{supportcubic}, \textit{adaptivecubic}, \textit{cubic}, \textit{gyroid}, \textit{3dhoneycomb}, \textit{honeycomb}, and \textit{concentric}. In the second row, the first five panels, from left to right, show \textit{line}, \textit{stars}, \textit{triangles}, \textit{grid}, and \textit{rectilinear}. The final five panels show layers~1--5 of the full Husky Head with the \textit{lightning} infill pattern.}

    \label{fig:husky_90_allvalid}
\end{figure}

\begin{figure}
\centering 
\includegraphics[width=1\linewidth]{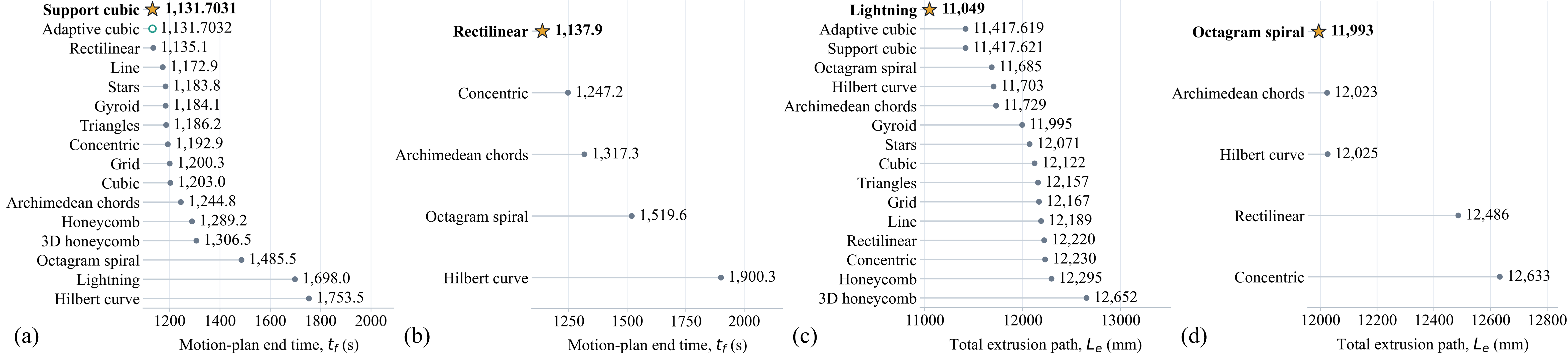} \caption{Quantitative infill-pattern screening for the Husky Head queries under 90\% and 100\% infill densities. Panels (a) and (b) rank realizable patterns by motion-plan end time for the time-minimization objective at 90\% and 100\% infill, respectively. Panels (c) and (d) rank realizable patterns by total extrusion path for the material-saving objective at 90\% and 100\% infill, respectively. Gold stars denote the selected patterns. In panel (a), the outlined marker denotes a rounded-value tie between \textit{adaptivecubic} and \textit{supportcubic}; \textit{supportcubic} is selected using the full-precision tool outputs. At 100\% infill, only realizable patterns are shown.} 
\label{fig:husky_pattern_screening} 
\end{figure}

At 90\% infill density, the Slicer Tool generates valid G-code for all 16 default infill patterns. Figure~\ref{fig:husky_90_allvalid} shows representative slicer-generated layers. The quantitative pattern-screening results are summarized in Fig.~\ref{fig:husky_pattern_screening}. For the time-minimization objective, panel~(a) ranks the realizable patterns by motion-plan end time. \textit{Supportcubic} is selected at $1131.7031\,\mathrm{s}$ using the full-precision outputs, practically tied with \textit{adaptivecubic} at $1131.7032\,\mathrm{s}$. Compared with \textit{hilbertcurve} at $1753.49\,\mathrm{s}$, the selected pattern reduces motion-plan end time by 35.5\%. For the material-saving objective, panel~(c) ranks the patterns by total extrusion path. \textit{Lightning} is selected because it gives the shortest extrusion path, $11049.11\,\mathrm{mm}$. Compared with \textit{3dhoneycomb}, the largest extrusion-path case at $12652.44\,\mathrm{mm}$, this corresponds to a 12.7\% reduction.

At 100\% infill density, the feasible pattern set becomes smaller under the current slicer configuration. Only \textit{rectilinear}, \textit{concentric}, \textit{hilbertcurve}, \textit{archimedeanchords}, and \textit{octagramspiral} generate valid G-code for this density and solid-fill configuration; the remaining default patterns fail slicer validation or are not realizable under this configuration and are removed before robot-side evaluation. For the time-minimization objective, Fig.~\ref{fig:husky_pattern_screening} (b) shows that \textit{rectilinear} gives the shortest motion-plan end time, $1137.91\,\mathrm{s}$, a 40.1\% reduction relative to \textit{hilbertcurve} at $1900.25\,\mathrm{s}$. For the material-saving objective, Fig.~\ref{fig:husky_pattern_screening} (d) shows that \textit{octagramspiral} gives the shortest total extrusion path, $11993.30\,\mathrm{mm}$, a 5.1\% reduction relative to \textit{concentric} at $12632.71\,\mathrm{mm}$.

After infill-pattern screening, orientation--placement selection is performed using perimeter-and-solid-infill evaluation G-code, generated by the Placement Specialist at each orientation--placement candidate with the infill density overwritten to 0\%. Setting the density to zero empties only the internal sparse-infill layers, chiefly the second layer, which becomes empty, while the other layers remain essentially unchanged, because their material is deposited as top/bottom solid-infill motion associated with the shell rather than as ordinary infill. The comparison therefore reflects the perimeter and solid-infill motion instead of the objective-dependent internal infill. The selected orientation--placement pair is used only after the final infill-containing G-code has been re-sliced and re-verified.

Table~\ref{tab:husky_placement_ranking} reports representative candidates from the lexicographic Joint-6 jerk ranking over the 15 no-infill orientation--placement cases. The selected orientation--placement pair is $90^\circ$/C3. Compared with the least favorable shown candidate, $45^\circ$/C1, the selected pair reduces $j_{6}^{\max}$ from $798.290\,\mathrm{rad\,s^{-3}}$ to $563.067\,\mathrm{rad\,s^{-3}}$, a 29.5\% reduction, and reduces $\bar{j}_{6}$ from $7.443\,\mathrm{rad\,s^{-3}}$ to $6.112\,\mathrm{rad\,s^{-3}}$, a 17.9\% reduction.

\begin{table}[t]
\centering
\caption{Representative Joint-6 jerk ranking for Husky Head orientation--placement selection under the perimeter-and-solid-infill evaluation setting.}
\label{tab:husky_placement_ranking}
\small
\begin{tabular}{c c c c c}
\hline
Rank & Orientation & Placement & $j_{6}^{\max}$ $(\mathrm{rad\,s^{-3}})$& $\bar{j}_{6}$ $(\mathrm{rad\,s^{-3}})$\\
\hline
1  & $90^\circ$ & C3 & 563.067 & 6.112 \\
2  & $60^\circ$ & C3 & 602.324 & 6.185 \\
3  & $0^\circ$  & C3 & 640.401 & 5.848 \\
4  & $90^\circ$ & C2 & 653.045 & 6.679 \\
5  & $45^\circ$ & C3 & 669.861 & 6.191 \\
\hline
11 & $60^\circ$ & C2 & 705.141 & 6.786 \\
12 & $60^\circ$ & C1 & 711.719 & 7.574 \\
13 & $0^\circ$  & C1 & 714.007 & 5.805 \\
14 & $90^\circ$ & C1 & 732.653 & 7.329 \\
15 & $45^\circ$ & C1 & 798.290 & 7.443 \\
\hline
\end{tabular}
\end{table}

\begin{figure}
    \centering
    \includegraphics[width=1\linewidth]{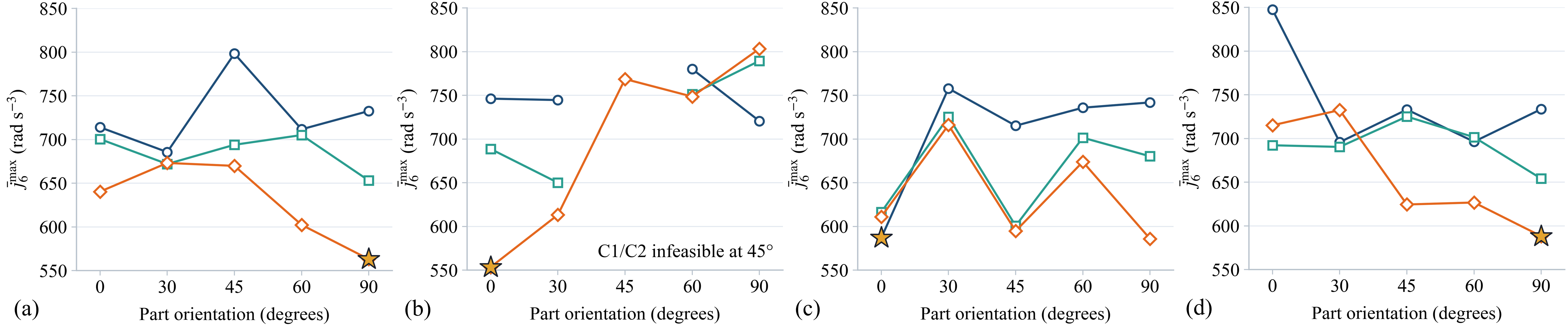}
    \caption{Orientation--placement interaction for maximum absolute Joint-6 jerk in (a) the five-layer Husky Head with infill set to 0\% while retaining the top and bottom solid layers; (b) the four-layer Gator Head under the perimeter-only profile; (c) the four-layer Husky Head under the perimeter-only profile; and (d) the five-layer Husky Head under the perimeter-only profile. The horizontal axis denotes part orientation, and the three connected profiles correspond to placements C1--C3. Gold stars mark the reported selected candidates, while gaps denote infeasible configurations. Connecting lines are used only to guide comparison across the tested discrete orientations and do not imply interpolation.}
    \label{fig:placement_heatmaps}
\end{figure}

\begin{figure}
    \centering
    \includegraphics[width=\textwidth]{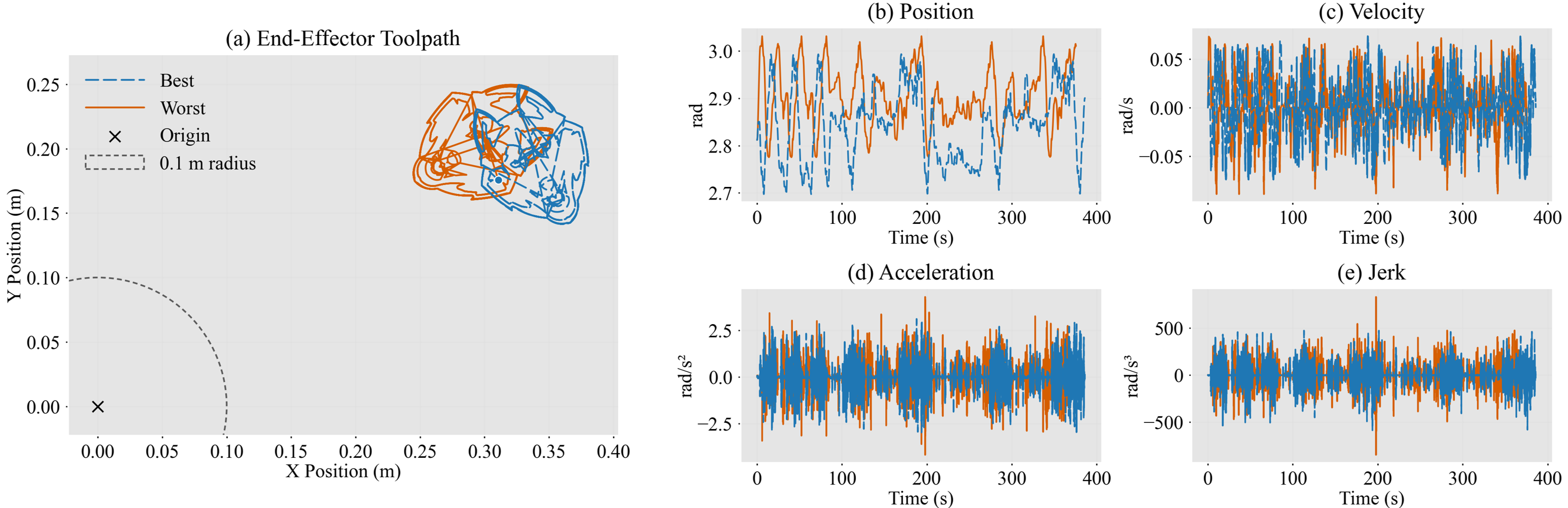}
    \caption{Comparison of the best and least favorable orientation--placement candidates in the Husky Head perimeter-and-solid-infill evaluation. The best candidate is $90^\circ$/C3, while the least favorable candidate is $45^\circ$/C1. (a) End-effector XY toolpaths. (b)--(e) Joint-6 position, velocity, acceleration, and jerk profiles, respectively.}
    \label{fig:husky_placement_best_worst_compare}
\end{figure}

The corresponding orientation--placement interaction is shown in Fig.~\ref{fig:placement_heatmaps} (a). Lower values indicate more favorable Joint-6 motion quality under the reported ranking rule. Figure~\ref{fig:husky_placement_best_worst_compare} compares the best and least favorable orientation--placement candidates. As shown in Fig.~\ref{fig:husky_placement_best_worst_compare} (a), changing the part orientation and placement alters the end-effector path within the robot workspace, even though both candidates are evaluated using the same perimeter-and-solid-infill evaluation geometry. These spatial differences produce distinct Joint-6 motion responses, as shown in Fig.~\ref{fig:husky_placement_best_worst_compare} (b)--(e), with the best candidate exhibiting lower jerk than the least favorable candidate.

\begin{figure}
    \centering
    \includegraphics[width=1\linewidth]{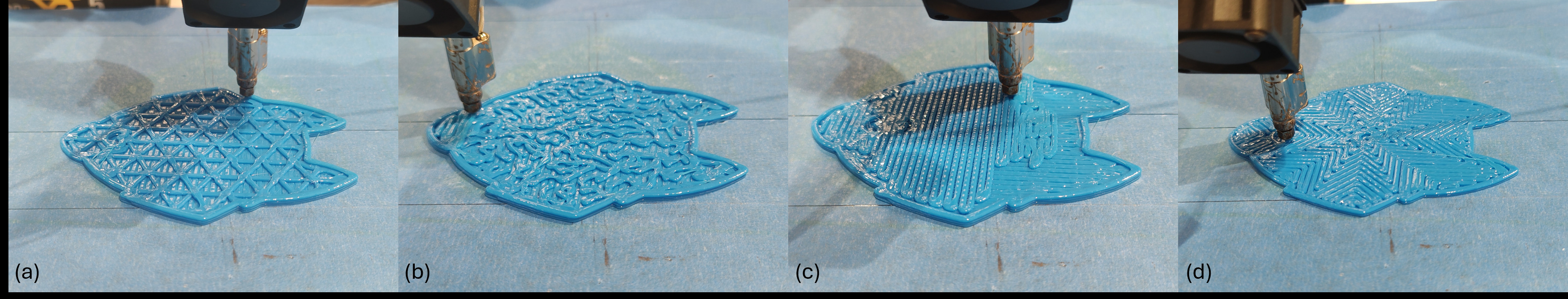}
    \caption{Printed second-layer infill for the selected Husky Head configurations: (a) fastest, 90\% infill, \textit{supportcubic}; (b) save filament, 90\% infill, \textit{lightning}; (c) fastest, 100\% infill, \textit{rectilinear}; and (d) save filament, 100\% infill, \textit{octagramspiral}.}
    \label{fig:husky_final_infill}
\end{figure}

\begin{figure}
    \centering
    \includegraphics[width=1\linewidth]{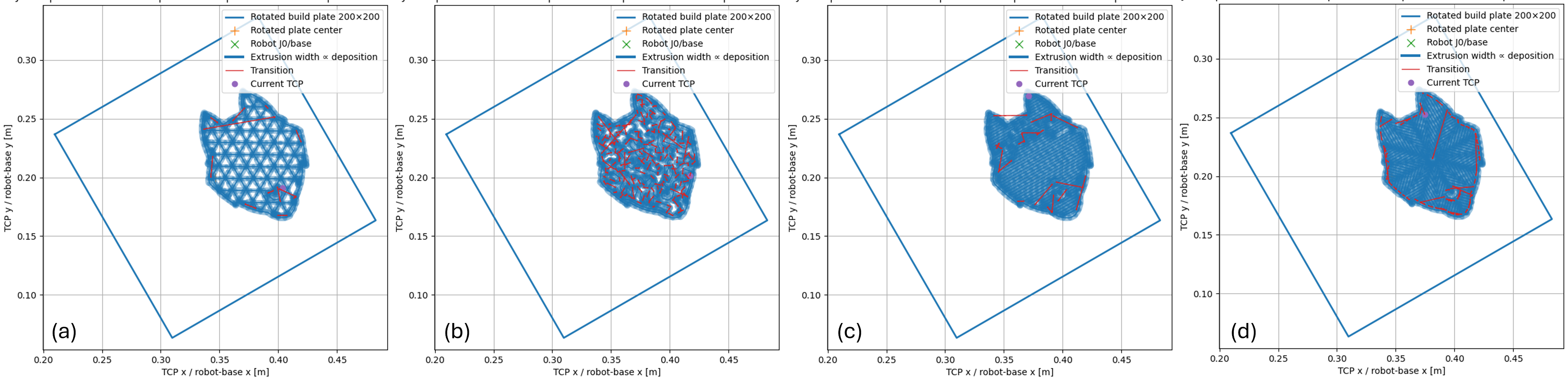}
    \caption{Layer-wise deposition visualization from the digital-shadow pipeline. Screenshots are taken at the end of the second-layer infill for (a) fastest, 90\% infill, \textit{supportcubic}; (b) save filament, 90\% infill, \textit{lightning}; (c) fastest, 100\% infill, \textit{rectilinear}; and (d) save filament, 100\% infill, \textit{octagramspiral}.}
    \label{fig:husky_digital_shadow}
\end{figure}

\begin{figure}
    \centering
    \includegraphics[width=\linewidth]{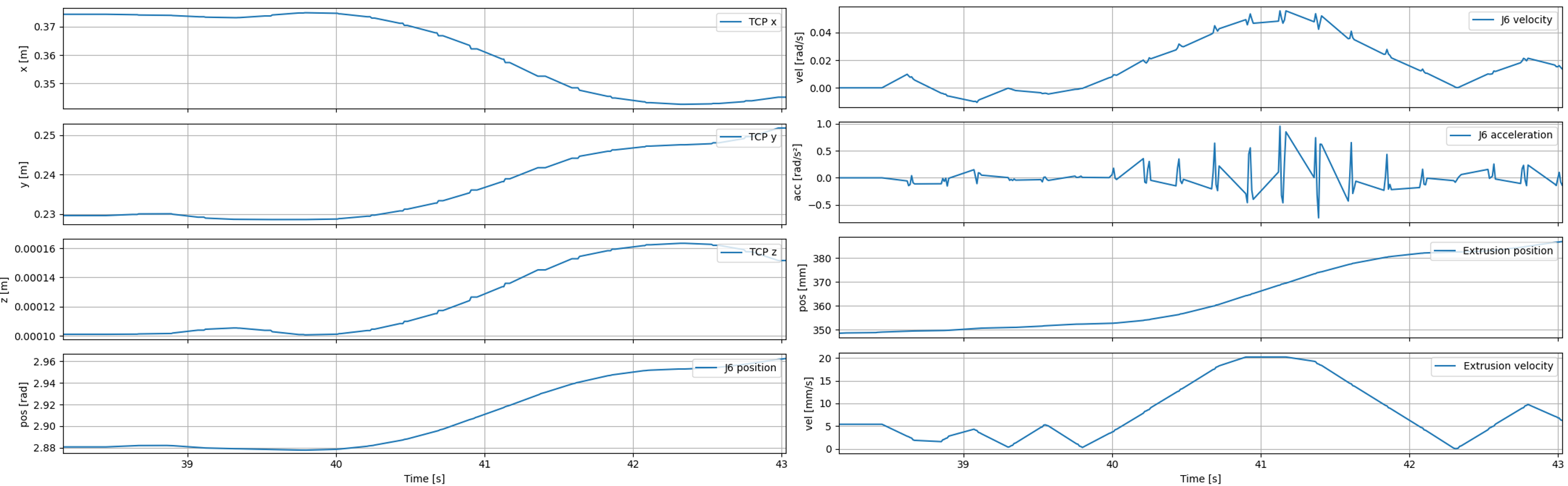}
    \caption{Execution-state monitoring from the digital-shadow pipeline during printing. The left column shows the commanded TCP position in $x$, $y$, and $z$, together with the corresponding Joint-6 position. The right column shows Joint-6 velocity and acceleration, as well as extrusion position and extrusion velocity. These synchronized signals provide a time-resolved view of robot motion and material-deposition states during execution.}
    \label{fig:husky_status}
\end{figure}

\begin{figure}
    \centering
    \includegraphics[width=\linewidth]{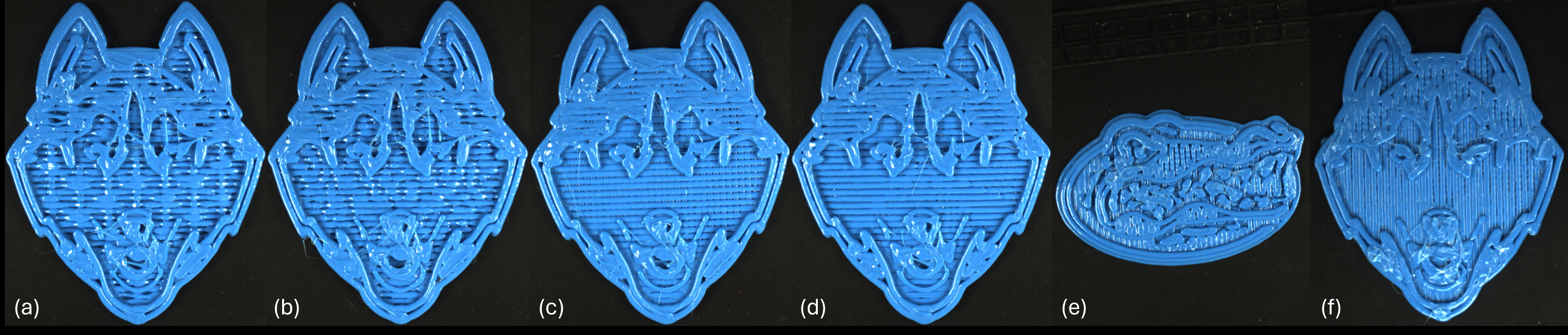}
    \caption{Physical prints of representative selected final plans: (a) Husky Head, fastest plan at 90\% infill using \textit{supportcubic}; (b) Husky Head, material-saving plan at 90\% infill using \textit{lightning}; (c) Husky Head, fastest plan at 100\% infill using \textit{rectilinear}; (d) Husky Head, material-saving plan at 100\% infill using \textit{octagramspiral}; (e) four-layer Gator Head; and (f) four-layer Husky Head.}
    \label{fig:physical_print_gallery}
\end{figure}

After orientation--placement selection, each final pattern--orientation--placement configuration is regenerated with the target infill density and re-verified by the Motion Planning Specialist using $\Delta t_{\mathrm{final}}=0.005\,\mathrm{s}$. The resulting plans are then printed with the digital-shadow pipeline activated. Figure~\ref{fig:husky_final_infill} shows the printed second-layer infill for the four selected configurations, Fig.~\ref{fig:husky_digital_shadow} and Fig.~\ref{fig:husky_status} provide the digital-shadow view of the printing process, including layer-wise command-side deposition visualization and synchronized monitoring of TCP motion, Joint-6 motion, and extrusion states. The corresponding physical prints are shown in Fig.~\ref{fig:physical_print_gallery} (a)--(d). Together, the results show that A-RAM applies objective-dependent pattern screening while retaining a shared robot-aware orientation--placement check and synchronized execution monitoring across the four density--objective combinations.

\subsection{Case Study III: Geometry-Dependent Orientation--Placement Selection}
\label{sec:case3_geometry}

This case study compares three parts: a four-layer Gator Head, a four-layer Husky Head, and a five-layer Husky Head. To isolate geometry-dependent orientation--placement effects, the infill pattern is fixed to \textit{rectilinear} for all three parts, while part orientation and workspace placement remain open search variables. The same controlled query is used for all three parts: “Use rectilinear infill at 100\% density and select the orientation and placement that prioritize perimeter motion quality only.” This query represents a perimeter-prioritized placement scenario motivated by the structural importance of the perimeter. According to PrusaSlicer, part strength is governed primarily by the number of perimeters rather than by infill, while internal infill mainly supports subsequent layers \cite{prusa_layers_perimeters}, and top and bottom solid infill primarily determines the external surface finish. Accordingly, unlike Case Study II, where both perimeter and top/bottom solid-infill trajectories are retained because the query also prioritizes visible solid-surface motion, the present query excludes both internal infill and top/bottom solid infill from the placement criterion.

The evaluation G-code therefore contains only perimeter trajectories. The prescribed 100\% infill density remains a final manufacturing requirement and is restored after orientation--placement selection; it does not define the motion-quality scope used to rank candidates. This contrast with Case Study II illustrates the query-conditioned evaluation behavior of the framework: the placement-evaluation trajectory is generated according to the motion-quality regions specified in the user request rather than from a fixed evaluation profile. Under this controlled setting, any change in the selected orientation--placement pair among the three parts is caused by part geometry and layer-stack differences rather than by changes in objective, infill pattern, or search workflow.

\begin{table}[t]
\centering
\caption{Agentic execution trace summary for the geometry-dependent orientation--placement tasks.}
\label{tab:case3_trace_summary}
\small
\setlength{\tabcolsep}{4pt}
\begin{tabular}{l l l l}
\hline
Part & Search scope & Placement-evaluation setting & Selected plan \\
\hline
Gator Head, 4-layer & $1\times5\times3=15$ & infill->0\%, without solid layer & \textit{rectilinear}, $0^\circ$, C3 \\
Husky Head, 4-layer & $1\times5\times3=15$ & infill->0\%, without solid layer & \textit{rectilinear}, $0^\circ$, C1 \\
Husky Head, 5-layer & $1\times5\times3=15$ & infill->0\%, without solid layer & \textit{rectilinear}, $90^\circ$, C3 \\
\hline
\end{tabular}
\end{table}

The Placement Tool generates the same five-orientation by three-placement candidate structure for each part under the build-plate constraints. Figure~\ref{fig:combined_candidate_generation} (c) shows the candidate-placement result for the four-layer Gator Head. The five-layer Husky Head placement result was already presented in Fig.~\ref{fig:combined_candidate_generation} (b); the four-layer Husky Head is not shown separately because its XY placement layout is visually the same as the five-layer version.

For placement evaluation, the Placement Specialist regenerates G-code for each candidate with infill density overwritten to 0\% and the top and bottom solid layers removed, so that the motion comparison reflects the perimeter-focused query. Table~\ref{tab:case3_best_worst_summary} summarizes the selected and least favorable candidates for the three parts under the perimeter-only placement-evaluation setting. For the four-layer Husky Head, $90^\circ$/C3 has a slightly lower raw $j_{6}^{\max}$ than $0^\circ$/C1 ($585.846$ versus $586.949\,\mathrm{rad\,s^{-3}}$); however, both candidates fall within the 1\% near-tie range, and the tolerance-aware lexicographic rule selects $0^\circ$/C1 because it has the lower $\bar{j}_{6}$ ($10.865$ versus $11.009\,\mathrm{rad\,s^{-3}}$). Relative to the least favorable candidate in each evaluated set, the selected candidate reduces $j_{6}^{\max}$ by 31.1\%, 22.6\%, and 30.6\% for the four-layer Gator Head, four-layer Husky Head, and five-layer Husky Head, respectively. The corresponding $\bar{j}_{6}$ values decrease by 5.6\% and 9.0\% for the four-layer Gator Head and four-layer Husky Head, respectively, whereas the five-layer Husky Head exhibits a 4.2\% increase in $\bar{j}_{6}$.

\begin{table}[t]
\centering
\caption{Selected and least favorable orientation--placement candidates for the geometry-dependent orientation--placement study under the perimeter-only evaluation setting.}
\label{tab:case3_best_worst_summary}
\small
\setlength{\tabcolsep}{4pt}
\renewcommand{\arraystretch}{1.1}
\begin{tabular}{l c c c c c}
\hline
Part & Case & Orientation & Placement & $j_{6}^{\max}$ $(\mathrm{rad\,s^{-3}})$& $\bar{j}_{6}$ $(\mathrm{rad\,s^{-3}})$ \\
\hline
Gator Head & Selected & $0^\circ$ & C3 & 553.46 & 8.82 \\
Gator Head & Least favorable & $90^\circ$ & C3 & 803.38 & 9.34 \\
\hline
Husky Head, 4-layer & Selected & $0^\circ$ & C1 & 586.95 & 10.87 \\
Husky Head, 4-layer & Least favorable & $30^\circ$ & C1 & 757.91 & 11.93 \\
\hline
Husky Head, 5-layer & Selected & $90^\circ$ & C3 & 588.20 & 11.19 \\
Husky Head, 5-layer & Least favorable & $0^\circ$ & C1 & 847.39 & 10.74 \\
\hline
\end{tabular}
\end{table}

The orientation--placement interactions in Fig.~\ref{fig:placement_heatmaps} (b)--(d) show that the preferred orientation--placement pair differs across geometries even under the same objective, workflow, pattern setting, and perimeter-only evaluation scope. The four-layer Gator Head selects $0^\circ$/C3, the four-layer Husky Head selects $0^\circ$/C1, and the five-layer Husky Head selects $90^\circ$/C3. The distinct interaction profiles indicate that the effect of placement varies with orientation and geometry, and that the preferred pair under the stated ranking rule differs across the evaluated geometries.

\begin{figure}
    \centering
    \includegraphics[width=1\linewidth]{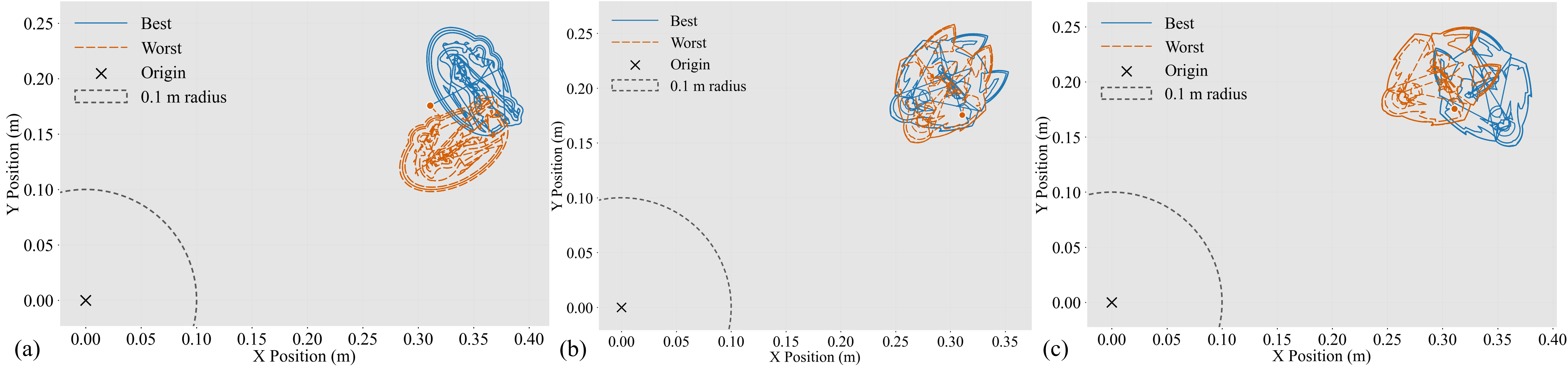}
    \caption{End-effector XY toolpaths of the selected and least favorable orientation--placement candidates for (a) four-layer Gator Head, (b) four-layer Husky Head, and (c) five-layer Husky Head under the perimeter-only evaluation setting. Blue trajectories denote the selected candidates and orange trajectories denote the least favorable candidates.}
    \label{fig:case3_placement_eecompare}
\end{figure}

Figure~\ref{fig:case3_placement_eecompare} compares the end-effector XY trajectories of the selected and least favorable candidates for the three parts. The trajectories occupy different locations and exhibit different path geometries relative to the robot base, leading to distinct robot-side motion behavior. The difference between the four-layer and five-layer Husky Head is especially important: under the prescribed tolerance-aware ranking rule, a small layer-stack change shifts the selected orientation--placement pair from $0^\circ$/C1 to $90^\circ$/C3.

After orientation--placement selection, the Slicer Tool regenerates the selected plans using the query-specified 100\% infill density, and the Motion Planning Specialist performs the final single-candidate rerun with $\Delta t_{\mathrm{final}}=0.005\,\mathrm{s}$. Physical realizations of the four-layer Gator Head and the four-layer Husky Head are shown in Fig.~\ref{fig:physical_print_gallery} (e)--(f). These results confirm that the preferred orientation--placement pair is geometry-dependent: the four-layer Gator and Husky Heads select different pairs because their geometries differ, while the four-layer and five-layer Husky Heads differ despite identical in-plane geometry, showing that even a small layer-stack change can reshape the Joint-6 jerk landscape and shift the selected orientation--placement pair.

\section{Conclusion}
\label{sec:conclusion}

This paper introduced agentic robotic additive manufacturing (A-RAM) for robotic-arm fused filament fabrication. A-RAM formulates robotic AM planning as a coupled decision problem involving slicer variables, part orientation, workspace placement, inverse kinematics, trajectory timing, motion quality, and extrusion. It converts a natural-language request and part STL into a structured planning workflow in which the LLM interprets the manufacturing objective and constraints, reasons over which planning variables are prescribed or require exploration, and formulates the structured planning intent; a deterministic Planning Agent instantiates that intent as the corresponding search and evaluation workflow, and domain tools compute the execution-critical evidence used for candidate selection and monitored execution.

The experimental results show that the selected slicer profile and orientation--placement pair vary with the specified objective and evaluated geometry. A-RAM changes its search scope according to the user objective and query specificity, while the selected plan also varies with part geometry. In the expert-specified Dogbone case, placement-only evaluation reduced the maximum and mean absolute Joint-6 jerk by 10.4\% and 18.3\%, respectively. In the goal-only Dogbone case, the 240-candidate search reduced these metrics by 53.5\% and 48.3\%, respectively, relative to the least favorable valid candidate. In the Husky Head study, objective-specific infill-pattern screening reduced motion-plan completion time by up to 40.1\% and extrusion path length by up to 12.7\%. The geometry-dependent placement study further showed that different parts, including layer-stack variants of similar geometries, select different orientation--placement pairs under the same objective and evaluation setting.

These findings show that robotic AM process planning must be query-specific, part-specific, and robot-aware, and that slicer-side quantities alone are insufficient for selecting a robotic manufacturing plan. By combining LLM-based manufacturing-intent and planning-variable reasoning with schema-constrained orchestration and deterministic tool-based verification, A-RAM provides a reproducible and auditable planning loop. Future work will extend the framework toward richer robot-dynamics metrics, closed-loop extrusion sensing, multi-objective optimization, and broader robotic AM processes.


\bibliographystyle{model1-num-names}
\bibliography{cas-refs}

\end{document}